\documentclass{article} %
\usepackage{iclr2027_conference,times}

\usepackage{amsmath,amsfonts,bm}

\def\eqref#1{equation~\ref{#1}}

\def\1{\bm{1}}

\DeclareMathAlphabet{\mathsfit}{\encodingdefault}{\sfdefault}{m}{sl}
\SetMathAlphabet{\mathsfit}{bold}{\encodingdefault}{\sfdefault}{bx}{n}

\usepackage{hyperref}
\usepackage{url}
\usepackage{booktabs}
\usepackage{array}
\usepackage{graphicx}
\usepackage{pifont}
\usepackage{float}
\usepackage[table]{xcolor}
\usepackage{comment}
\usepackage{wrapfig}
\usepackage{capt-of}
\usepackage{subcaption}
\hypersetup{
    colorlinks=true,      %
    citecolor=blue,       %
    linkcolor=black,      %
    urlcolor=black        %
}

\title{\raisebox{-0.1\height}{\includegraphics[width=0.5cm]{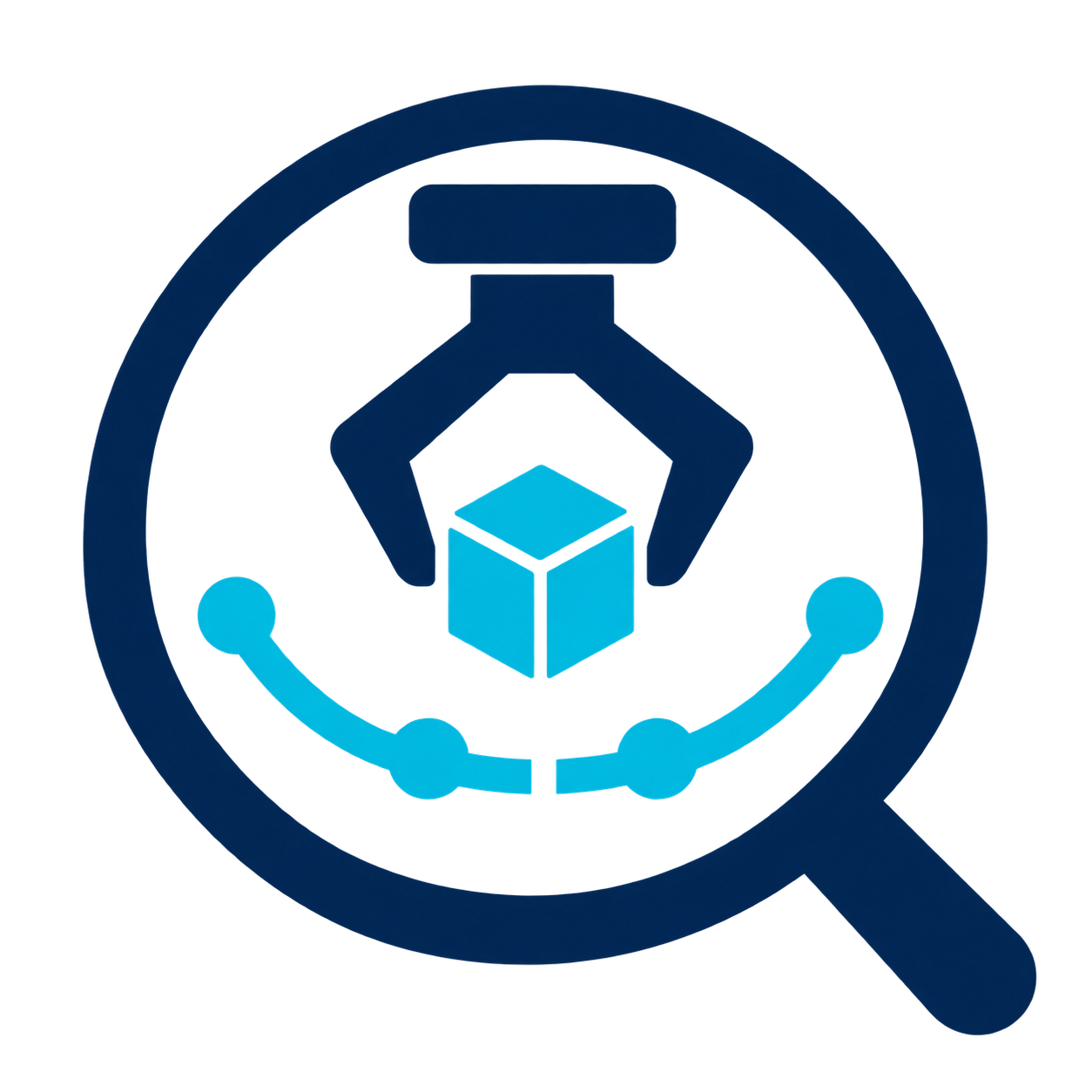}}~PhysicsLENS: Diagnosing Physical Property Blindness in Video Generation Models}

\author{Isaiah Milkey, Som Sagar, Aditya Taparia, \\
\textbf{Xinyuan Liu, Jiqing Wen, Ransalu Senanayake} \\
School of Computing and Augmented Intelligence \\
Arizona State University \\
Tempe, AZ, USA \\
\texttt{\{imilkey,ssagar6,ataparia,xinyua11,jwen31,rsenana1\}@asu.edu}
}
\definecolor{Everyone}{RGB}{0, 150, 136}  %
\definecolor{Rans}{RGB}{228, 26, 28}      %
\definecolor{Jiqing}{RGB}{152, 78, 163} %
\definecolor{Xinyuan}{RGB}{77, 175, 74}   %
\definecolor{Aditya}{RGB}{255, 127, 0}    %
\definecolor{Som}{RGB}{166, 86, 40}       %
\definecolor{Isaiah}{RGB}{172, 145, 21}    %

\iclrfinalcopy %
\begin{document}

\maketitle
\lhead{Preprint.}

\begin{abstract}

Reliable video world models could provide scalable predictive environments for robot learning, planning, and evaluation. However, generated robot videos can violate physical principles and complete tasks through physically implausible behavior, limiting their reliability for robot learning and planning. 
Current video-generation benchmarks exclude physics that are inherently hidden by visuals (e.g., weight, viscosity, friction). Due to this, video models are evaluated on the fidelity of physics, not the underlying accuracy of physics. 
We introduce PhysicsLENS, a dataset and benchmark for evaluating plausibility of physical properties grounded in robotics. PhysicsLENS uses matched scenario pairs that hold the same conditioning frame and task, while varying underlying physics in the scene description. Scenarios are curated from public robot video sources and annotated across seven physical domains: collision, gravity, momentum, friction, deformation, fluid, and causality. We evaluate across four video generation models, producing over 400 human-annotated labels. 
Results show that plausible-looking videos often ignore the stated property (34 of 47), and that stating the property lowers plausibility only slightly and not significantly.

\end{abstract}

\section{Introduction}
Video generation models could provide scalable predictive environments for robot learning \citep{unisim2024,cosmoswfm2025}, planning \citep{du2023unipi}, and evaluation \citep{quevedo2025worldgym}. By generating diverse interactions without repeatedly collecting real-world data, these models could expose robots to rare events, unfamiliar objects, and changing physical conditions \citep{dreamgen2025}. However, increasing the volume of generated data is useful only if the videos reliably represent the physical consequences of actions. A video may appear realistic and depict successful task completion while violating friction, momentum, object permanence, or other physical constraints. Such violations undermine the usefulness of generated videos as predictive environments, yet may go undetected by benchmarks focused on visual realism and task success. We therefore require a benchmark that evaluates whether generated interactions respect the physical properties and constraints that govern them.

Existing video-generation benchmarks capture this reliability only partially. Most benchmarks focus on visual quality, temporal consistency, or motion plausibility. These criteria can identify obvious artifacts, but cannot determine whether a model correctly grounds the physical properties specified in its conditioning input. Properties such as weight, friction, viscosity, and deformability may not be observable from a single image. Consequently, visually identical scenes can require different outcomes: a robot should interact differently with a heavy object than with a light one, or with a slippery surface than with a rough one. A model can therefore generate a plausible video while producing behavior that is physically incorrect for the stated condition.

\begin{figure*}
  \centering
  \includegraphics[width=0.9\textwidth]{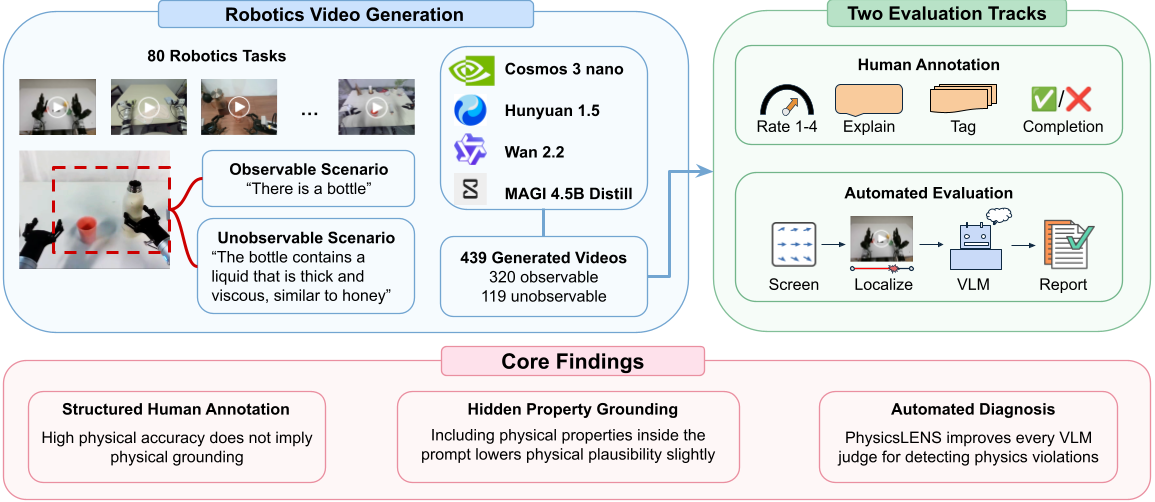}
  \vspace{-0.5em}
  \caption{Overview of PhysicsLENS: robotics video generation, evaluation, and core findings.}
  \label{fig:physicslens_overview}
\end{figure*}

We introduce PhysicsLENS, a benchmark for evaluating both physical plausibility and physical-property grounding in robot video generation. The dataset comprises of curated image--text prompts, more than 400 clips generated by four video generation models, and structured human annotations for every clip. We construct prompts from public robot videos, grouping scenarios into seven physical domains: collision, gravity, momentum, friction, deformation, fluid, and causality. In unobservable scenarios, the text specifies a physical property that cannot be inferred from that frame. Matched unobservable prompts keep the visual context and task fixed while varying the specified property, allowing us to test whether generated interactions reflect the corresponding physical consequences. Figure~\ref{fig:physicslens_overview} summarizes the benchmark workflow.

The benchmark evaluates generated clips through two complementary tracks. In the human evaluation track, annotators rate physical plausibility and task completion, describe and categorize visible physical violations, and assess whether interactions in unobservable scenarios reflect the stated property. In the automated evaluation track, a four-stage hierarchical pipeline screens each clip, localizes potential violations, verifies them through targeted vision-language model questions, and reports supporting evidence.  Across four video-generation models and 439 generated clips, the results show that a video's apparent physical plausibility does not guarantee correct grounding in a hidden property. Our contributions are:
\begin{enumerate}
\item We introduce PhysicsLENS, a robotics-focused benchmark covering seven physical domains and distinguishing physical plausibility from physical-property grounding.
\item We construct matched scenarios that preserve visual context while varying text-specified physical conditions, including properties that cannot be inferred from appearance alone.
\item We present a multi-stage evaluation procedure and benchmark four video-generation models with more than 400 human judgments, finding that plausible-looking videos often ignore the stated property.
\end{enumerate}

\section{Related Work}

\newcommand{\lensyes}{\textcolor{green!60!black}{\ding{51}}}
\newcommand{\lensno}{\textcolor{red!70!black}{\ding{55}}}

\paragraph{Evaluating physics in generated video.}
Video world models are increasingly used as simulators and data sources for robot learning \citep{unisim2024,cosmoswfm2025,dreamgen2025}, yet videos that look realistic can still break physics \citep{kang2025howfar,motamed2026physicsiq}. Distributional metrics and general benchmark suites mainly measure perceptual quality \citep{unterthiner2018towards,huang2024vbench}, although VBench-2.0 adds physics dimensions \citep{zheng2025vbench}. Physics benchmarks score commonsense physics in generated videos \citep{bansal2025videophy,bansal2026videophy,meng2024phygenbench,gu2026phyworldbenchcomprehensiveevaluationphysical}, compare generations with real experiments or recorded outcomes \citep{zhang2025morpheus,motamed2026physicsiq}, or turn measured motion into verifiable rewards \citep{le2025gravity}. Physion-Eval and PQSG also localize failures, using expert reasoning traces and VLM-generated question graphs, respectively \citep{zhang2026physionevalevaluatingphysicalrealism,pothiraj2026physicsquestionscenegraph}. For robotics, WorldSimBench, EWMBench, RBench, and RoboWM-Bench evaluate generated manipulation videos on visual quality, motion, task adherence, or whether the motion can be executed as actions \citep{qin2025worldsimbench,yue2025ewmbench,RBench_deng2026rethinkingvideogenerationmodel,jiang2026robowmbenchbenchmarkevaluatingworld}. WorldModelBench scores instruction following and physics adherence across domains that include robotics \citep{li2025worldmodelbenchjudgingvideogeneration}, and RoboGaze reports temporally localized glitches under a robotics-specific taxonomy \citep{nguyen2026robogaze}. To our knowledge, none of these benchmarks separates properties that are visible in the conditioning frame from properties stated only in the prompt (Table~\ref{tab:related}), so a high score cannot show whether a model followed the stated physics or only what the scene looks like.
 
\paragraph{Hidden physical properties.}
Inferring properties that appearance does not reveal is a classic test of physical understanding: Physion++ and ComPhy ask models to infer mass, friction, elasticity, or charge from observed interactions \citep{tung2023physionevaluatingphysicalscene,comphy2022}. PhysicsLENS reverses this flow of information: the property is stated in text, and the generator must turn it into motion. Controllable generators receive physical quantities as dedicated inputs, such as forces or mass and friction coefficients \citep{physgen2024,forceprompting2025,phyparam2026}. T2V-CompBench tests whether prompt attributes bind to the right objects, but only for visible attributes such as color, shape, and motion direction \citep{sun2025t2vcompbenchcomprehensivebenchmarkcompositional}. We instead test whether general-purpose image-to-video models follow a property given only in plain language, with the frame and task held fixed.

\begin{table*}[t]
\centering
\scriptsize
\setlength{\tabcolsep}{5pt}
\renewcommand{\arraystretch}{1}
\caption{
Comparison of physics benchmarks and video critics.
Ref.-free requires no paired real-world recording;
Obs.\ axis tests prompt-specified properties not inferable
from the initial frame.
}
\label{tab:related}

\begin{tabular}{@{}l l *{6}{c}@{}}
    \toprule
    Category
    & Method
    & Physics
    & Robotics
    & Human
    & Ref.-free
    & Localized
    & Obs.\ axis \\
    \midrule

    Physics benchmarks
    & VideoPhy-2~\citep{bansal2026videophy}
    & \lensyes & \lensno & \lensyes & \lensyes & \lensno & \lensno \\
    & PhyGenBench~\citep{meng2024phygenbench}
    & \lensyes & \lensno & \lensno & \lensyes & \lensno & \lensno \\
    & Morpheus~\citep{zhang2025morpheus}
    & \lensyes & \lensno & \lensno & \lensno & \lensno & \lensno \\
    & Physics-IQ~\citep{motamed2026physicsiq}
    & \lensyes & \lensno & \lensno & \lensno & \lensno & \lensno \\
    & NewtonRewards~\citep{le2025gravity}
    & \lensyes & \lensno & \lensno & \lensyes & \lensno & \lensno \\
    & GeoPhys~\citep{internò2026geophysgeometryphysicalplausibility}
    & \lensyes & \lensno & \lensno & \lensyes & \lensyes & \lensno \\

    \midrule

    Robotics evaluators
    & RoboGaze~\citep{nguyen2026robogaze}
    & \lensyes & \lensyes & \lensyes & \lensyes & \lensyes & \lensno \\
    & RBench~\citep{RBench_deng2026rethinkingvideogenerationmodel}
    & \lensyes & \lensyes & \lensyes & \lensyes & \lensno & \lensno \\
    & RoboWM-Bench~\citep{jiang2026robowmbenchbenchmarkevaluatingworld}
    & \lensyes & \lensyes & \lensno & \lensno & \lensyes & \lensno \\

    \midrule

    Latent properties
    & Physion++~\citep{tung2023physionevaluatingphysicalscene}
    & \lensyes & \lensno & \lensyes & \lensno & \lensno & \lensno \\
    & T2V-CompBench~\citep{sun2025t2vcompbenchcomprehensivebenchmarkcompositional}
    & \lensno & \lensno & \lensno & \lensyes & \lensyes & \lensno \\

    \midrule

    VLM reasoning
    & Cosmos-Reason1~\citep{nvidia2025cosmosreason}
    & \lensyes & \lensyes & \lensno & \lensyes & \lensyes & \lensno \\
    & QuantiPhy~\citep{puyin2026quantiphy}
    & \lensyes & \lensno & \lensyes & \lensyes & \lensno & \lensno \\
    & MASS~\citep{wu2026mass}
    & \lensyes & \lensno & \lensyes & \lensyes & \lensyes & \lensno \\

    \midrule

    \rowcolor{orange!15}
    \textbf{This Work}
    & \textbf{PhysicsLENS}
    & \lensyes & \lensyes & \lensyes
    & \lensyes & \lensyes & \lensyes \\
    \bottomrule
\end{tabular}

\end{table*}

\paragraph{VLM judges of physical plausibility.}
\label{sec:VLM-judges}
Automatic physics evaluation often relies on vision-language models (VLMs), either fine-tuned or prompted \citep{bansal2025videophy,meng2024phygenbench,nvidia2025cosmosreason}, but their reliability is limited. VLMs struggle with physical-world understanding \citep{physbench2025}, most models score near chance on intuitive-physics tests \citep{intphys2_2025}, and plausible verbal explanations can hide wrong estimates of velocity and acceleration \citep{puyin2026quantiphy}. Multimodal judges also show biases and inconsistent judgments \citep{mllmjudge2024}, and video VLMs have limited fine-grained motion and temporal grounding \citep{hong2025motionbench,wang2025grounded}. Measured evidence helps: MASS adds depth, object grounding, and motion tracking to VLM reasoning \citep{wu2026mass}, and GeoPhys finds that VLM judges detect physics violations near chance while a training-free score on frozen image-encoder features does better \citep{internò2026geophysgeometryphysicalplausibility}. PhysicsLENS follows this direction by combining tracked measurements with targeted VLM questions, and we test it directly against human annotations.

\section{PhysicsLENS Benchmark}

\subsection{Observable and Unobservable Conditions}
\label{sec:observability}

\begin{figure}[t]
\centering
\includegraphics[width=0.85\linewidth]{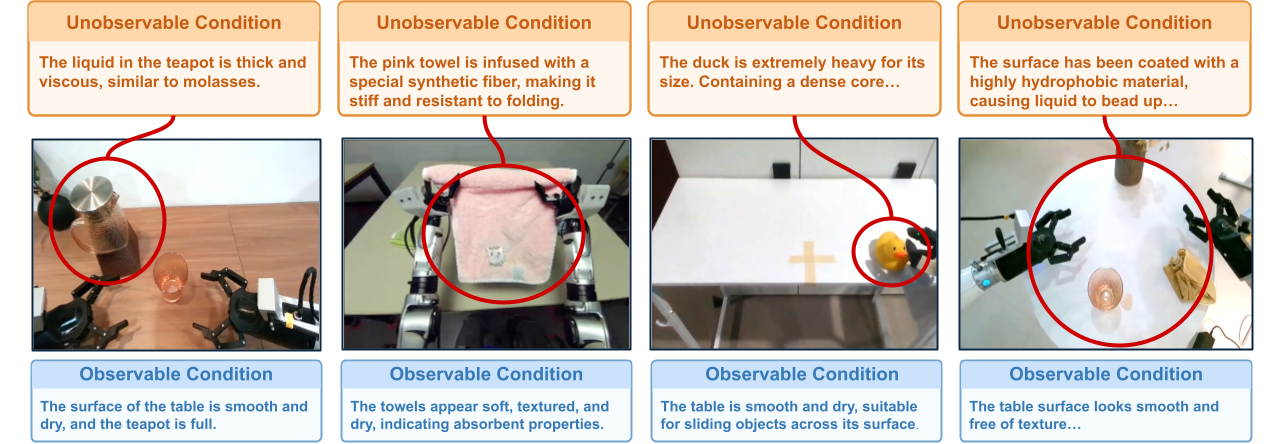}
\vspace{-0.6em}
\caption{Example matched prompt pairs. Each pair shares the initial frame and task. The observable description (blue) states what the frame shows; the unobservable description (orange) replaces it with a latent property that departs from what the frame suggests.}
\label{fig:matched-prompt-construction}
\end{figure}

Image-conditioned video generation relies on two sources of information: an initial frame and a text prompt describing the scene and requested task. In ordinary robot videos, visual appearance is often correlated with physical properties; for example, size may correlate with mass, and visible wetness may suggest low surface friction. A model may therefore generate plausible motion from visual priors without grounding the physical properties specified in text. PhysicsLENS evaluates whether models follow text-specified properties when those properties cannot be inferred from the initial frame. We define two conditions. In the \emph{observable} condition, the prompt is consistent with the physical properties suggested by the frame. In the \emph{unobservable} condition, the same frame is paired with a prompt that specifies a latent property that departs from those visual expectations. For example, a prompt may state that liquid inside an ordinary teapot is ``thick and viscous, similar to molasses'' (Figure~\ref{fig:matched-prompt-construction}). A model that relies only on appearance may instead generate water-like flow. We refer to this failure to condition on a text-specified latent property as \emph{physical property blindness}.

\paragraph{Matched-pair construction.} Each unobservable scenario is derived from an observable counterpart with the same initial frame and task instruction. Only the scene description changes: the observable description is consistent with the frame, whereas the unobservable description introduces a latent property that conflicts with its visual appearance. This construction holds the visual scene, task, and frame quality fixed, making the textual property the sole intervention. Because we generate only one video per prompt and model, individual pairwise differences may reflect generation noise; we therefore analyze matched pairs in aggregate. The unobservable scenarios specify four task-relevant properties: viscosity for pouring (10 scenarios), surface condition for wiping (8), elasticity for deformable objects (7), and mass for pushing (5). Each property is expressed in plain language within the scene description provided to the video generator. The generator receives neither a reference continuation nor the expected outcome.

The design supports two complementary evaluations. \emph{Property grounding} assesses whether the generated video visibly reflects the stated latent property, using the human adherence score (Section~\ref{sec:human-annotation}). \emph{Matched-condition comparison} tests whether specifying that property affects physical plausibility while holding the initial frame and task fixed—that is, whether failure to express the property coincides with visibly less plausible physics. Together, these evaluations distinguish general visual plausibility from adherence to the specified physical condition.

\subsection{Scenarios and Generated Videos}
\label{sec:dataset-design}

PhysicsLENS contains 110 prompts: 80 observable scenarios and 30 matched unobservable variants. The 80 starting frames come from 11 public robot datasets that cover humanoids, dual-arm and single-arm manipulators, and mobile manipulators (Table~\ref{tab:sources}). Each frame shows the scene before the robot interacts with solid objects or liquids. Appendices~\ref{app:dataset-details} and~\ref{app:annotation-details} describe frame preparation, prompt generation, and the annotation schema. We used GPT-4o to draft structured scene and action descriptions from each frame and its logged task. An author subsequently reviewed the descriptions.

We generate one video per prompt using each of four models: Wan 2.2 \citep{wan2025wan}, Cosmos 3 Nano~\citep{agarwal2026cosmos}, HunyuanVideo 1.5~\citep{wu2025hunyuanvideo}, and MAGI 4.5B Distill~\citep{teng2025magi}. Together, they span diverse generative paradigms, including diffusion video models, autoregressive video generation, and world/action foundation modeling. One unobservable HunyuanVideo 1.5 prompt did not produce a usable video. The realized benchmark hence contains 439 videos: 320 observable and 119 unobservable. HunyuanVideo 1.5 contributes 109 videos, including 29 unobservable videos, while each other model contributes 110. Since the benchmark contains one generation per prompt--model combination, the reported results characterize these generated clips and do not estimate variability across generation seeds.

\subsection{Human Annotation}
\label{sec:human-annotation}

Annotators score each video on up to three scales, which we report separately because they can disagree. \emph{Physical plausibility} $P\in\{1,2,3,4\}$ judges only the visible physics: contact, motion, deformation, and object persistence. \emph{Task completion} $A\in\{0,1\}$ records whether the task is done by the end, even if the physics looks wrong. \emph{Hidden-property adherence} $H\in\{1,2,3,4\}$, rated only for unobservable videos, records whether the outcome shows the stated property. We treat $P\geq3$ as mostly or fully plausible and $H\geq3$ as mostly or fully following the property.

When $P<4$, annotators must describe the physics violation in the video and select one or more of eight violation labels. The labels are the seven physical domains plus contact, which annotators could select separately from collision; we merge contact into collision in all analyses. Appendix~\ref{app:annotation-details} gives the full rubric, the label definitions, and how videos were assigned to annotators.

\subsection{Automated Diagnostic Pipeline}
\label{sec:pipeline}

\begin{figure}[t]
    \centering
    \includegraphics[width=\textwidth]{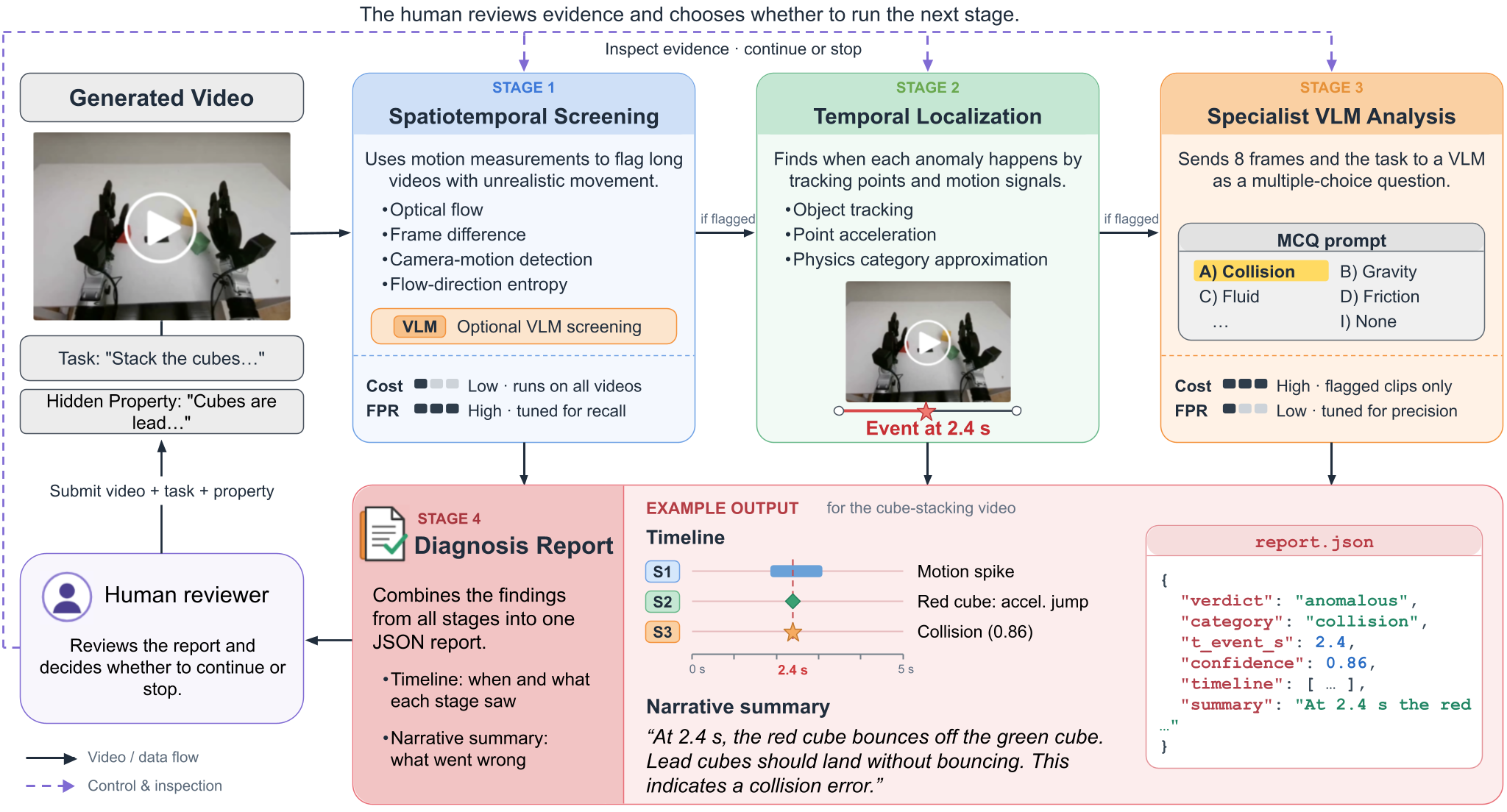}
    \vspace{-1em}
\captionof{figure}{Evidence flow through the PhysicsLENS diagnostic pipeline.}
\label{fig:pipeline-evidence-flow}
\end{figure}

PhysicsLENS also includes an automated evaluator (Figure~\ref{fig:pipeline-evidence-flow}). Because VLM judges are unreliable on their own (Section~\ref{sec:VLM-judges}), the evaluator pairs motion measurements, which record how objects and the camera move, with targeted VLM questions, which identify what went wrong. Its input is the generated video and the task. For unobservable scenarios, the hidden-property question also receives the stated property and the scenario's expected outcome. Human labels are used only to fit how the evidence is combined, on training folds that exclude the scenario being judged.

The evaluator has four stages.
\emph{Screening} measures motion over the first 5 seconds of the video: frame differences, optical flow, camera motion, flow-direction entropy, and how DINOv2 frame embeddings change. It also asks a VLM a physics-error question that lists specific errors instead of asking about overall realism.
\emph{Localization} tracks points with camera motion removed and measures their kinematics: speed, acceleration, the time of peak acceleration, and points that disappear inside the frame rather than at its border, the sign of an object vanishing mid-video.
\emph{VLM verification} shows each VLM eight frames and the task and asks three questions: (i)~whether the task was completed; (ii)~which of eight failure descriptions (one per physical domain, plus an object appearing or vanishing) best describes the main physics problem, if any; and (iii)~for unobservable scenarios, whether the outcome shows the stated property.
\emph{Reporting} merges the evidence into a per-video JSON report with each finding's time window, object, violation type, confidence, and a short explanation. We do not evaluate time localization because our human labels are not time-dependent. The released tool also runs per-domain \emph{specialists} in Stage~3, which our experiments do not use (Appendix~\ref{app:pipeline-details}).

To compare with human ratings (Section~\ref{sec:results-tool}), we pose these questions to ten VLMs and read each answer from the probabilities of its first output token. For each question, we rank each VLM's scores across videos, average the ranks over the ten VLMs, and mix the result with five screening and localization signals using a single weight. The five signals and the weight are selected on the training folds. Reports are not scored, because the human labels have no time stamps.

\section{Experimental Results}
\label{sec:results}

This section presents results on the PhysicsLENS benchmark and evaluates its diagnostic pipeline through four research questions. RQ1--RQ3 are answered with human annotations, and RQ4 compares the pipeline against them. \textbf{RQ1:} Does high physical plausibility imply that a model followed a grounded specification (Sec.~\ref{sec:results-grounding})? \textbf{RQ2:} Does stating a latent property in text change plausibility (Sec.~\ref{sec:results-paired})? \textbf{RQ3:} What do individual failure cases reveal about the distinction between grounding and plausibility (Sec.~\ref{sec:results-cases})? \textbf{RQ4:} Can the PhysicsLENS pipeline approximate human judgments (Sec.~\ref{sec:results-tool})?

\subsection{RQ1: Does High Plausibility Imply Property Grounding?}
\label{sec:results-grounding}

A video may look physically plausible without reflecting the physical property specified in its prompt. To test this, we collected annotations rating each video's plausibility ($P$) and its adherence to the stated property ($H$) on 1--4 scales (Sec.~\ref{sec:human-annotation}). We consider a video plausible when $P \geq 3$ and property-adherent when $H \geq 3$.

Across the total of 119 videos with unobservable properties, human annotators noted 47 as plausible. Yet 34 of those 47 (72\%) do not adhere to the stated property ($H \leq 2$). This is the \emph{physical property blindness} described in Section~\ref{sec:observability}. The pattern appears for all four generators (Table~\ref{tab:hidden-results}). MAGI has the highest mean plausibility but the lowest mean adherence (Fig.~\ref{fig:joint-ratings}); it accounts for 21 of the 34 plausible but ungrounded videos. The result persists without MAGI: 13 of the remaining 26 plausible videos (50\%) ignore the property.

\newsavebox{\PLtab}\newsavebox{\PLcap}
\begin{figure}[t]
    \centering
    \sbox{\PLtab}{\begin{minipage}[t]{0.52\linewidth}
        \vspace{0pt}
        \centering
        \setlength{\abovecaptionskip}{0pt}
        \captionof{table}{Human evaluation on unobservable scenarios. $\bar{P}$, $\bar{H}$: mean plausibility and adherence (1--4). Compl.: completion rate.}
        \label{tab:hidden-results}
        \footnotesize
        \setlength{\tabcolsep}{2pt}
        \resizebox{\linewidth}{!}{%
        \begin{tabular}{@{}l r r r r r@{}}
            \toprule
            Model & $n$ & $\bar{P}$ & $\bar{H}$ & Compl. & $H{\le}2 \mid P{\ge}3$ \\
            \midrule
            HunyuanVideo 1.5  & 29 & 2.24 & 2.24 & 72.4\% & 5/13 (38.5\%) \\
            MAGI 4.5B Distill & 30 & 3.10 & 1.10 &  6.7\% & 21/21 (100\%) \\
            Wan 2.2           & 30 & 1.53 & 1.90 & 40.0\% & 2/5 (40.0\%) \\
            Cosmos 3 Nano     & 30 & 2.07 & 1.87 & 66.7\% & 6/8 (75.0\%) \\
            \bottomrule
        \end{tabular}}
    \end{minipage}}%
    \sbox{\PLcap}{\begin{minipage}[t]{0.45\linewidth}
        \vspace{0pt}
        \setlength{\abovecaptionskip}{4pt}
        \captionof{figure}{Mean plausibility ($P$) vs.\ mean adherence ($H$) per model (1--4).}
        \label{fig:joint-ratings}
    \end{minipage}}%
    \usebox{\PLtab}\hfill
    \begin{minipage}[t]{0.45\linewidth}
        \vspace{0pt}
        \centering
        \includegraphics[width=\linewidth,
            height=\dimexpr\dp\PLtab-\dp\PLcap\relax]{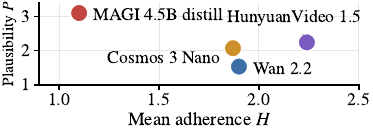}\par\nointerlineskip
        \usebox{\PLcap}
    \end{minipage}
\end{figure}

Conversely, of the 28 videos that follow the property ($H \geq 3$), 15 are rated implausible ($P \leq 2$). These videos reflect the stated property but violate other aspects of physics. \textbf{Thus, plausibility and property grounding must be evaluated separately: a video can satisfy either criterion without satisfying the other.}

\subsection{RQ2: Does Stating A Latent Property In Text Change Plausibility?}
\label{sec:results-paired}

We next compare how the same physical property is presented. Each unobservable scenario has an observable counterpart with the same frame, task, and generator. Only the description differs: it either matches the frame or specifies a different latent property (e.g., molasses-like tea in an ordinary teapot). This comparison tests whether the presentation of the property changes the plausibility of the generated video.

\begin{figure}[t]
    \centering
    \sbox{\PLtab}{\begin{minipage}[t]{0.485\linewidth}
        \vspace{0pt}
        \centering
        \setlength{\abovecaptionskip}{0pt}
        \captionof{table}{Matched plausibility comparison. Obs., Unobs.: mean $P$ on the matched pairs; $\Delta$: unobservable minus observable. $p$: Holm-adjusted paired test; pooled row: scenario-level sign-flip test over the 29 scenarios annotated for all four models.}
        \label{tab:paired-results}
        \footnotesize
        \setlength{\tabcolsep}{3pt}
        \resizebox{\linewidth}{!}{%
        \begin{tabular}{@{}lrrrrr@{}}
            \toprule
            Model & $n$ & Obs. & Unobs. & $\Delta$ & $p$ \\
            \midrule
            HunyuanVideo 1.5  & 29 & 2.62 & 2.24 & $-0.38$ & 0.709 \\
            MAGI 4.5B Distill & 30 & 3.33 & 3.10 & $-0.23$ & 0.884 \\
            Wan 2.2           & 30 & 1.83 & 1.53 & $-0.30$ & 0.497 \\
            Cosmos 3 Nano     & 30 & 2.20 & 2.07 & $-0.13$ & 0.884 \\
            \midrule
            Pooled            & 29 & ---  & ---  & $-0.28$ & 0.085 \\
            \bottomrule
        \end{tabular}}
    \end{minipage}}%
    \sbox{\PLcap}{\begin{minipage}[t]{0.485\linewidth}
        \vspace{0pt}
        \setlength{\abovecaptionskip}{4pt}
        \captionof{figure}{Paired change $\Delta = P^{\mathrm{unobs}} - P^{\mathrm{obs}}$ per model and pooled, with 95\% bootstrap intervals over scenarios.}
        \label{fig:paired-effects}
    \end{minipage}}%
    \usebox{\PLtab}\hfill
    \begin{minipage}[t]{0.485\linewidth}
        \vspace{0pt}
        \centering
        \includegraphics[width=\linewidth,
            height=\dimexpr\dp\PLtab-\dp\PLcap\relax]{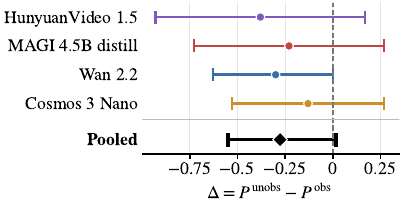}\par\nointerlineskip
        \usebox{\PLcap}
    \end{minipage}
\end{figure}

Mean plausibility is lower in the unobservable condition for all four generators, by 0.13 to 0.38 points on the 1--4 scale (Table~\ref{tab:paired-results}). Across individual pairs, plausibility decreases in 40 cases and increases in 26. Pooling the 29 scenarios generated by all four models gives a mean decrease of 0.28 points (Fig.~\ref{fig:paired-effects}; $p=0.085$). \textbf{Stating a property that the frame does not show lowers plausibility for all four generators, but the drop is small and not statistically significant.}

\subsection{RQ3: What Do Failure Cases Reveal About Grounding and Plausibility?}
\label{sec:results-cases}

\begin{figure}[t]
    \centering
    \vspace{-0.7em}
    \includegraphics[width=\linewidth]{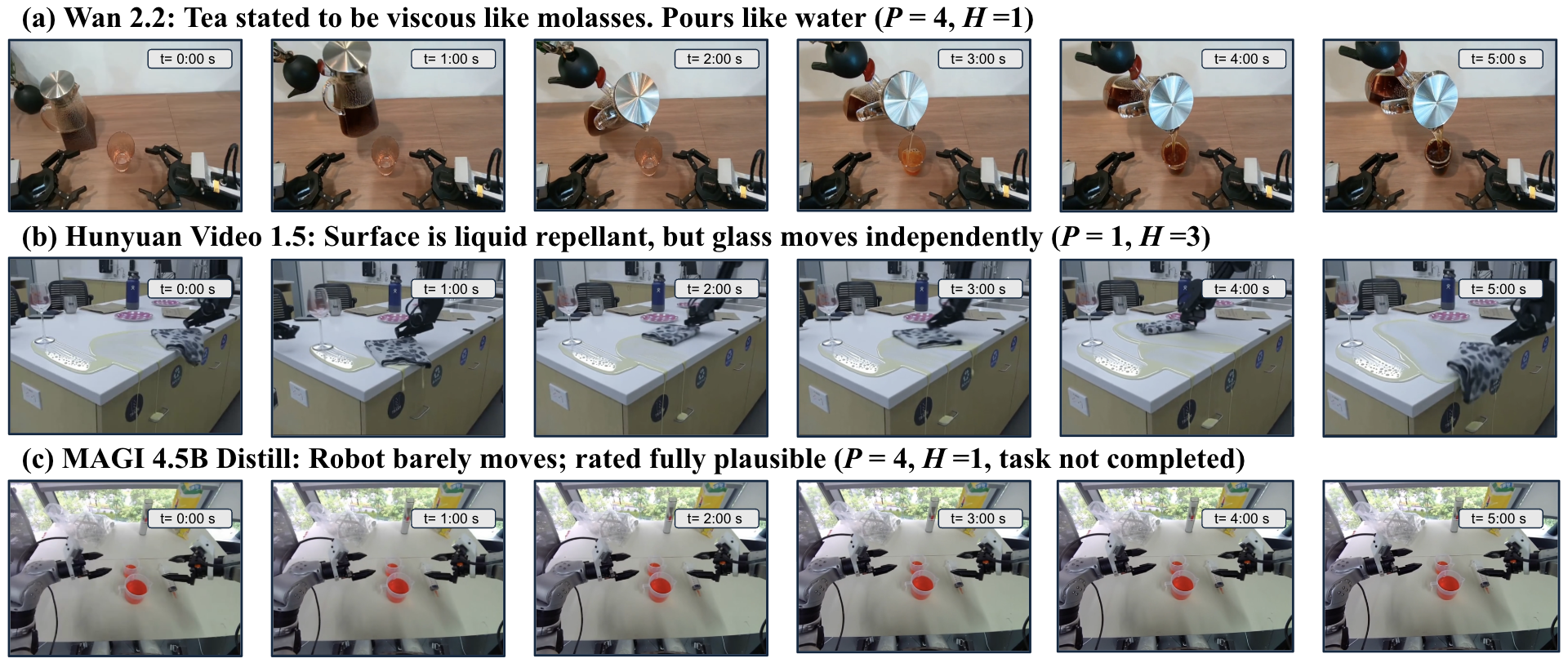}
    \vspace{-0.7em}
    \caption{Three annotated failure cases (six frames each). $P$: physical plausibility; $H$: hidden-property adherence (both 1--4). (a) A clean, realistic pour that ignores the stated viscosity. (b) A video rated as following the stated property that is nevertheless physically implausible. (c) A nearly static video rated fully plausible.}
    \label{fig:failure-cases}
\end{figure}

Fig.~\ref{fig:failure-cases} illustrates why a plausibility rating alone cannot establish property grounding. In a task that specifies molasses-like tea, all four generators produce plausible pours that fail to reflect its viscosity ($H{=}1$). Wan~2.2, for example, pours the tea like water while receiving the highest plausibility rating (Fig.~\ref{fig:failure-cases}a). A different case arises when the requested interaction barely occurs. MAGI's robot often remains nearly still (Fig.~\ref{fig:failure-cases}c), producing a video that looks plausible without demonstrating how the stated property affects the task.

On the contrary, a video can reflect the stated property while violating other physical constraints. With a silicone-coated, liquid-repellent table, HunyuanVideo~1.5 receives an adherence rating of $H{=}3$, but the glass moves without contact, making the video implausible (Fig.~\ref{fig:failure-cases}b). In another case, Wan~2.2 leaves an extremely heavy tape dispenser stationary ($H{=}4$), but the robot's hand passes through it ($P{=}1$). PhysicsLENS ranks the visible violation in Fig.~\ref{fig:failure-cases}b among the benchmark's least plausible videos. \textbf{These cases show why physical plausibility and adherence to the stated property require separate evaluations.}

\begin{table*}[t]
    \centering
    \footnotesize
    \setlength{\tabcolsep}{2.6pt}
    \caption{PhysicsLENS against human judgment (AUC on 439 videos; 0.5 is chance). \emph{Generator identity} scores each video by its generator's average rating; \emph{signals} are the PhysicsLENS Stage-1/2 measurements; \emph{debiased} is the PhysicsLENS physics-error question; the \emph{ensemble} averages all ten VLMs. Values are mean $\pm$ standard deviation across the three runs, except for $^\dagger$ rows, which report mean $\pm$ standard deviation across the ten VLMs. Bold indicates the best result in each column.}
    \label{tab:main-results}
    \begin{tabular}{@{}lccccccc@{}}
        \toprule
        & \multicolumn{3}{c}{Physical plausibility} & Task & Violation & Physics & Hidden \\
        \cmidrule(lr){2-4}
        System & Obs. & Unobs. & All & completion & detection & family & property \\
        \midrule
        Generator identity & 0.63{\scriptsize\textcolor{gray}{$\pm$0.03}} & \textbf{0.72{\scriptsize\textcolor{gray}{$\pm$0.05}}} & 0.65{\scriptsize\textcolor{gray}{$\pm$0.03}} & 0.72{\scriptsize\textcolor{gray}{$\pm$0.02}} & \textbf{0.73{\scriptsize\textcolor{gray}{$\pm$0.03}}} & 0.51{\scriptsize\textcolor{gray}{$\pm$0.02}} & 0.68{\scriptsize\textcolor{gray}{$\pm$0.05}} \\
        Signals only & 0.61{\scriptsize\textcolor{gray}{$\pm$0.03}} & 0.65{\scriptsize\textcolor{gray}{$\pm$0.05}} & 0.62{\scriptsize\textcolor{gray}{$\pm$0.03}} & 0.70{\scriptsize\textcolor{gray}{$\pm$0.03}} & 0.70{\scriptsize\textcolor{gray}{$\pm$0.03}} & 0.51{\scriptsize\textcolor{gray}{$\pm$0.02}} & 0.69{\scriptsize\textcolor{gray}{$\pm$0.05}} \\
        \midrule
        VLM$^\dagger$ & 0.53{\scriptsize\textcolor{gray}{$\pm$0.06}} & 0.53{\scriptsize\textcolor{gray}{$\pm$0.06}} & 0.53{\scriptsize\textcolor{gray}{$\pm$0.05}} & 0.71{\scriptsize\textcolor{gray}{$\pm$0.04}} & 0.58{\scriptsize\textcolor{gray}{$\pm$0.08}} & 0.59{\scriptsize\textcolor{gray}{$\pm$0.02}} & 0.57{\scriptsize\textcolor{gray}{$\pm$0.03}} \\
        VLM + signals$^\dagger$ & 0.60{\scriptsize\textcolor{gray}{$\pm$0.02}} & 0.66{\scriptsize\textcolor{gray}{$\pm$0.01}} & 0.61{\scriptsize\textcolor{gray}{$\pm$0.01}} & 0.76{\scriptsize\textcolor{gray}{$\pm$0.02}} & 0.72{\scriptsize\textcolor{gray}{$\pm$0.01}} & 0.58{\scriptsize\textcolor{gray}{$\pm$0.02}} & 0.70{\scriptsize\textcolor{gray}{$\pm$0.02}} \\
        Debiased VLM$^\dagger$ & 0.64{\scriptsize\textcolor{gray}{$\pm$0.07}} & 0.63{\scriptsize\textcolor{gray}{$\pm$0.04}} & 0.63{\scriptsize\textcolor{gray}{$\pm$0.06}} & 0.71{\scriptsize\textcolor{gray}{$\pm$0.04}} & 0.58{\scriptsize\textcolor{gray}{$\pm$0.08}} & 0.59{\scriptsize\textcolor{gray}{$\pm$0.02}} & 0.57{\scriptsize\textcolor{gray}{$\pm$0.03}} \\
        Debiased + signals$^\dagger$ & 0.65{\scriptsize\textcolor{gray}{$\pm$0.05}} & 0.67{\scriptsize\textcolor{gray}{$\pm$0.03}} & 0.65{\scriptsize\textcolor{gray}{$\pm$0.04}} & 0.76{\scriptsize\textcolor{gray}{$\pm$0.02}} & 0.72{\scriptsize\textcolor{gray}{$\pm$0.01}} & 0.58{\scriptsize\textcolor{gray}{$\pm$0.02}} & 0.70{\scriptsize\textcolor{gray}{$\pm$0.02}} \\
        \midrule
        PhysicsLENS ensemble & \textbf{0.73{\scriptsize\textcolor{gray}{$\pm$0.03}}} & 0.71{\scriptsize\textcolor{gray}{$\pm$0.05}} & \textbf{0.72{\scriptsize\textcolor{gray}{$\pm$0.02}}} & 0.77{\scriptsize\textcolor{gray}{$\pm$0.02}} & 0.67{\scriptsize\textcolor{gray}{$\pm$0.03}} & \textbf{0.63{\scriptsize\textcolor{gray}{$\pm$0.01}}} & 0.59{\scriptsize\textcolor{gray}{$\pm$0.06}} \\
        \rowcolor{orange!15}PhysicsLENS (full) & \textbf{0.73{\scriptsize\textcolor{gray}{$\pm$0.03}}} & 0.71{\scriptsize\textcolor{gray}{$\pm$0.05}} & \textbf{0.72{\scriptsize\textcolor{gray}{$\pm$0.02}}} & \textbf{0.81{\scriptsize\textcolor{gray}{$\pm$0.02}}} & \textbf{0.73{\scriptsize\textcolor{gray}{$\pm$0.03}}} & 0.62{\scriptsize\textcolor{gray}{$\pm$0.02}} & \textbf{0.71{\scriptsize\textcolor{gray}{$\pm$0.05}}} \\
        \bottomrule
    \end{tabular}
\end{table*}

\begin{figure*}[t]
    \centering
    \begin{minipage}[t]{0.48\textwidth}
        \centering
        \vspace{0pt}
        \includegraphics[width=\linewidth]{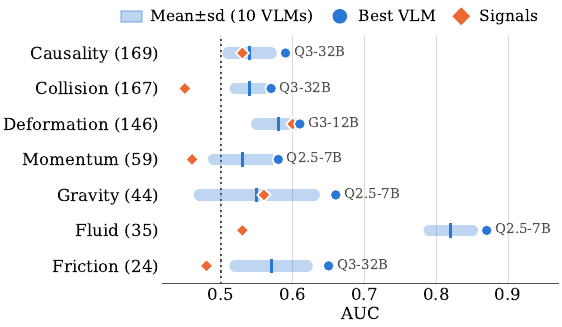}
        \vspace{-2em}
        \caption{Attribution accuracy by physics family: AUC for videos with this family vs. videos with a different violation (dotted line is chance).
        }
        \label{fig:category-attribution}
    \end{minipage}\hfill
    \begin{minipage}[t]{0.48\textwidth}
        \centering
        \vspace{0pt}
        \includegraphics[width=\linewidth]{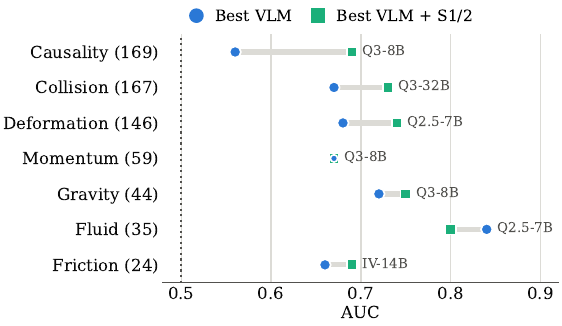}
        \vspace{-2em}
        \caption{Detection accuracy by physics family: AUC for videos with this family vs. the 106 videos with no violation (dotted line is chance). 
        }
        \label{fig:category-detection}
    \end{minipage}
\end{figure*}

\subsection{RQ4: How Well Do Automated Evaluators Match Human Judgments on PhysicsLENS?}
\label{sec:results-tool}

PhysicsLENS provides human annotations for 439 videos, allowing us to measure how well automated scores agree with human judgments. We evaluate ten open-weight VLMs from seven families, Stage-1/2 video signals, and combinations of the two. Combination weights and signal selection are fitted using five-fold cross-validation, with all versions of a scenario kept in the same fold. We report AUC, where 0.5 is chance and 1 is perfect ranking. 

\textbf{Combining VLM judgments with video signals improves agreement.}
The full evaluator combines answers from an ensemble of ten VLMs with Stage-1/2 signals. It reaches an AUC of 0.72 for overall plausibility, 0.81 for task completion, 0.73 for violation detection, and 0.71 for hidden-property adherence (Table~\ref{tab:main-results}). It achieves the highest or tied-highest score on five of the seven measures. Adding signals to an individual VLM raises mean overall plausibility AUC from 0.53 to 0.61, violation detection from 0.58 to 0.72, task completion from 0.71 to 0.76, and property adherence from 0.57 to 0.70. Signals alone reach 0.70 for violation detection; one frequently selected signal tracks objects vanishing during a video.

Generator identity is a strong baseline because the four generators differ substantially in quality. It slightly exceeds the full evaluator on unobservable-video plausibility (0.72 versus 0.71), but assigns the same score to every video from a given generator. Its within-generator AUC is therefore 0.5 by construction, whereas the automated evaluator reaches 0.67 for unobservable plausibility and 0.76 for task completion. Generator identity also cannot score videos from a new generator without first obtaining human ratings for that generator.

\textbf{Targeted questions improve VLM plausibility judgments.}
When asked directly whether a video is physically plausible, many VLMs perform near chance (Table~\ref{tab:backbones-obs-unobs}). To focus their judgments on physical behavior, we instead ask about specific errors, such as objects passing through one another, vanishing, floating, changing shape impossibly, or moving without contact. The question also instructs the VLMs to disregard task success and the amount of motion. It improves plausibility AUC for eight of the ten VLMs. On observable videos, Qwen3-VL-32B improves from 0.55 to 0.75 and InternVL3-14B from 0.51 to 0.71. On VideoPhy-2~\citep{bansal2026videophy} (Appendix~\ref{app:videophy}), Qwen3-VL-32B improves from 0.645 to 0.720.

\textbf{Hidden-property adherence remains difficult to judge.}
With the targeted physics-error question, Qwen3-VL-32B reaches plausibility AUCs of 0.75 on observable videos and 0.76 on unobservable videos. Its AUC for judging whether a text-specified property was followed, however, is only 0.57. Across the ten VLMs, adding signals raises mean adherence AUC from 0.57 to 0.70, close to signals alone (0.69) and generator identity (0.68). This agreement does not establish that the automated evaluator can determine whether a particular hidden property was realized.

\textbf{Violation detection is easier than attribution.}
The best VLM's detection AUC ranges from 0.56 to 0.84 across seven violation families; adding video signals improves five families, leaves momentum unchanged, and lowers performance for fluids (Figure~\ref{fig:category-detection}). Among videos with different violations, mean VLM attribution AUC ranges from 0.53 to 0.58 for most families, although fluids reach 0.82 (Figure~\ref{fig:category-attribution}). Since 66\% of videos with violations contain multiple types, assigning a single family is particularly difficult. \textbf{Automated scores agree with human judgments on several visible outcomes, but hidden-property grounding and precise violation attribution remain difficult.}

\begin{table}[t]
    \centering
    \footnotesize
    \setlength{\tabcolsep}{3pt}
    \caption{Plausibility AUC per VLM on observable ($n{=}320$) and unobservable ($n{=}119$) videos, with a standard question, the PhysicsLENS physics-error question, and that question plus Stage-1/2 signals. \emph{Hidden property}: whether the stated property was followed. Mean $\pm$ sd over three runs that show the VLM different frames of each video. Bold: best per column.}
    \label{tab:backbones-obs-unobs}
    \begin{tabular}{@{}lccccccc@{}}
        \toprule
        & \multicolumn{2}{c}{Standard} & \multicolumn{2}{c}{Physics-error} & \multicolumn{2}{c}{+ signals} & Hidden \\
        \cmidrule(lr){2-3}\cmidrule(lr){4-5}\cmidrule(lr){6-7}\cmidrule(l){8-8}
        VLM & Obs. & Unobs. & Obs. & Unobs. & Obs. & Unobs. & property \\
        \midrule
        Qwen3-VL-32B & 0.55{\scriptsize\textcolor{gray}{$\pm$0.00}} & 0.57{\scriptsize\textcolor{gray}{$\pm$0.02}} & \textbf{0.75{\scriptsize\textcolor{gray}{$\pm$0.00}}} & \textbf{0.76{\scriptsize\textcolor{gray}{$\pm$0.05}}} & \textbf{0.74{\scriptsize\textcolor{gray}{$\pm$0.00}}} & \textbf{0.76{\scriptsize\textcolor{gray}{$\pm$0.05}}} & 0.57{\scriptsize\textcolor{gray}{$\pm$0.01}} \\
        InternVL3-14B & 0.51{\scriptsize\textcolor{gray}{$\pm$0.01}} & 0.45{\scriptsize\textcolor{gray}{$\pm$0.02}} & 0.71{\scriptsize\textcolor{gray}{$\pm$0.01}} & 0.65{\scriptsize\textcolor{gray}{$\pm$0.00}} & 0.71{\scriptsize\textcolor{gray}{$\pm$0.02}} & 0.68{\scriptsize\textcolor{gray}{$\pm$0.01}} & 0.51{\scriptsize\textcolor{gray}{$\pm$0.01}} \\
        Qwen3-VL-8B & 0.54{\scriptsize\textcolor{gray}{$\pm$0.01}} & 0.58{\scriptsize\textcolor{gray}{$\pm$0.01}} & 0.68{\scriptsize\textcolor{gray}{$\pm$0.00}} & 0.70{\scriptsize\textcolor{gray}{$\pm$0.01}} & 0.68{\scriptsize\textcolor{gray}{$\pm$0.01}} & 0.71{\scriptsize\textcolor{gray}{$\pm$0.01}} & 0.56{\scriptsize\textcolor{gray}{$\pm$0.01}} \\
        Qwen2.5-VL-32B & 0.45{\scriptsize\textcolor{gray}{$\pm$0.01}} & 0.44{\scriptsize\textcolor{gray}{$\pm$0.01}} & 0.66{\scriptsize\textcolor{gray}{$\pm$0.02}} & 0.64{\scriptsize\textcolor{gray}{$\pm$0.02}} & 0.67{\scriptsize\textcolor{gray}{$\pm$0.01}} & 0.71{\scriptsize\textcolor{gray}{$\pm$0.01}} & 0.57{\scriptsize\textcolor{gray}{$\pm$0.01}} \\
        InternVL3-8B & \textbf{0.63{\scriptsize\textcolor{gray}{$\pm$0.00}}} & \textbf{0.62{\scriptsize\textcolor{gray}{$\pm$0.02}}} & 0.65{\scriptsize\textcolor{gray}{$\pm$0.01}} & 0.59{\scriptsize\textcolor{gray}{$\pm$0.01}} & 0.65{\scriptsize\textcolor{gray}{$\pm$0.01}} & 0.62{\scriptsize\textcolor{gray}{$\pm$0.01}} & 0.60{\scriptsize\textcolor{gray}{$\pm$0.01}} \\
        Llama-4-Scout-17B & 0.57{\scriptsize\textcolor{gray}{$\pm$0.01}} & 0.50{\scriptsize\textcolor{gray}{$\pm$0.02}} & 0.63{\scriptsize\textcolor{gray}{$\pm$0.02}} & 0.59{\scriptsize\textcolor{gray}{$\pm$0.04}} & 0.63{\scriptsize\textcolor{gray}{$\pm$0.01}} & 0.65{\scriptsize\textcolor{gray}{$\pm$0.01}} & 0.60{\scriptsize\textcolor{gray}{$\pm$0.02}} \\
        Qwen2.5-VL-7B & 0.46{\scriptsize\textcolor{gray}{$\pm$0.02}} & 0.53{\scriptsize\textcolor{gray}{$\pm$0.01}} & 0.60{\scriptsize\textcolor{gray}{$\pm$0.02}} & 0.61{\scriptsize\textcolor{gray}{$\pm$0.03}} & 0.61{\scriptsize\textcolor{gray}{$\pm$0.02}} & 0.66{\scriptsize\textcolor{gray}{$\pm$0.01}} & 0.56{\scriptsize\textcolor{gray}{$\pm$0.01}} \\
        Gemma-3-12B & 0.61{\scriptsize\textcolor{gray}{$\pm$0.01}} & \textbf{0.62{\scriptsize\textcolor{gray}{$\pm$0.02}}} & 0.59{\scriptsize\textcolor{gray}{$\pm$0.00}} & 0.60{\scriptsize\textcolor{gray}{$\pm$0.01}} & 0.60{\scriptsize\textcolor{gray}{$\pm$0.00}} & 0.65{\scriptsize\textcolor{gray}{$\pm$0.01}} & 0.59{\scriptsize\textcolor{gray}{$\pm$0.03}} \\
        LLaVA-OV-7B & 0.47{\scriptsize\textcolor{gray}{$\pm$0.00}} & 0.55{\scriptsize\textcolor{gray}{$\pm$0.01}} & 0.55{\scriptsize\textcolor{gray}{$\pm$0.01}} & 0.62{\scriptsize\textcolor{gray}{$\pm$0.00}} & 0.59{\scriptsize\textcolor{gray}{$\pm$0.00}} & 0.66{\scriptsize\textcolor{gray}{$\pm$0.01}} & 0.62{\scriptsize\textcolor{gray}{$\pm$0.00}} \\
        Mistral-Small-3.1-24B & 0.53{\scriptsize\textcolor{gray}{$\pm$0.01}} & 0.54{\scriptsize\textcolor{gray}{$\pm$0.01}} & 0.53{\scriptsize\textcolor{gray}{$\pm$0.00}} & 0.58{\scriptsize\textcolor{gray}{$\pm$0.02}} & 0.58{\scriptsize\textcolor{gray}{$\pm$0.01}} & 0.66{\scriptsize\textcolor{gray}{$\pm$0.01}} & 0.56{\scriptsize\textcolor{gray}{$\pm$0.01}} \\
        \midrule
        Signals only & --- & --- & --- & --- & 0.60{\scriptsize\textcolor{gray}{$\pm$0.01}} & 0.65{\scriptsize\textcolor{gray}{$\pm$0.00}} & \textbf{0.69{\scriptsize\textcolor{gray}{$\pm$0.05}}} \\
        \bottomrule
    \end{tabular}
\end{table}

\section{Conclusion}
We introduced PhysicsLENS, a benchmark for testing whether generated robot videos actually reflect the physical properties they are told to have. By keeping the initial frame and action the
same and only changing how the property is described, matched scenarios let us look at property grounding and physical plausibility side by side. Across 439 clips from four generators, human evaluation found a clear gap between the two: 34 of 47 clips rated plausible under unobservable conditions did not follow the stated property, and 15 of 28 clips that did follow it were
still physically implausible. In other words, a video can look right without actually respecting the physics it was asked to show, or get the property right while breaking something else
entirely. Pairing targeted VLM questions with measured video signals meaningfully improves agreement with these human judgments, reaching an AUC of 0.72 for plausibility and 0.81 for
task completion, though judging whether a specific hidden property was followed, and figuring out which physics family a violation belongs to, are both still hard problems. Taken together, these
results make the case for evaluating plausibility and property grounding as two separate questions, not one.

We kept this study deliberately focused on 80 initial frames and four video-generation models so we could isolate property grounding under tightly matched conditions. From here, a natural next step is testing more models, more hidden-property types, and multiple generations per prompt, which would help separate how much of the variation comes from the generation process itself versus the property being tested.

\section*{Acknowledgments}
We thank Anil Murthy, Eren Sadikoglu, Kacy Hatfield, Riana Chatterjee, Son Nguyen, Sreevishakh Vasudevan, and Vivek Sahukar for early discussions that helped shape this work.

\bibliography{iclr2026_conference}

\begin{thebibliography}{61}
\providecommand{\natexlab}[1]{#1}
\providecommand{\url}[1]{\texttt{#1}}
\expandafter\ifx\csname urlstyle\endcsname\relax
  \providecommand{\doi}[1]{doi: #1}\else
  \providecommand{\doi}{doi: \begingroup \urlstyle{rm}\Url}\fi

\bibitem[Agarwal et~al.(2026)Agarwal, Ali, Allen, Antolini, Aubame, Azzolini, Bai, Bala, Balaji, Bapst, et~al.]{agarwal2026cosmos}
Niket Agarwal, Arslan Ali, Jon Allen, Martin Antolini, Adeline Aubame, Alisson Azzolini, Junjie Bai, Maciej Bala, Yogesh Balaji, Josh Bapst, et~al.
\newblock {Cosmos 3}: Omnimodal world models for physical {AI}.
\newblock \emph{arXiv preprint arXiv:2606.02800}, 2026.

\bibitem[Bansal et~al.(2025)Bansal, Lin, Xie, Zong, Yarom, Bitton, Jiang, Sun, Chang, and Grover]{bansal2025videophy}
Hritik Bansal, Zongyu Lin, Tianyi Xie, Zeshun Zong, Michal Yarom, Yonatan Bitton, Chenfanfu Jiang, Yizhou Sun, Kai-Wei Chang, and Aditya Grover.
\newblock {VideoPhy}: Evaluating physical commonsense for video generation.
\newblock In \emph{International Conference on Learning Representations}, volume 2025, pp.\  102075--102121, 2025.

\bibitem[Bansal et~al.(2026)Bansal, Peng, Bitton, Goldenberg, Grover, and Chang]{bansal2026videophy}
Hritik Bansal, Clark Peng, Yonatan Bitton, Roman Goldenberg, Aditya Grover, and Kai-Wei Chang.
\newblock {VideoPhy-2}: A challenging action-centric physical commonsense evaluation in video generation.
\newblock In \emph{ICLR}, 2026.

\bibitem[Bordes et~al.(2025)Bordes, Garrido, Kao, Williams, Rabbat, and Dupoux]{intphys2_2025}
Florian Bordes, Quentin Garrido, Justine~T Kao, Adina Williams, Michael Rabbat, and Emmanuel Dupoux.
\newblock {IntPhys 2}: Benchmarking intuitive physics understanding in complex synthetic environments.
\newblock \emph{arXiv preprint arXiv:2506.09849}, 2025.

\bibitem[Bu et~al.(2025)Bu, Cai, Chen, Cui, Ding, Feng, Gao, He, Hu, Huang, et~al.]{agibotworld}
Qingwen Bu, Jisong Cai, Li~Chen, Xiuqi Cui, Yan Ding, Siyuan Feng, Shenyuan Gao, Xindong He, Xuan Hu, Xu~Huang, et~al.
\newblock {AgiBot World Colosseo}: A large-scale manipulation platform for scalable and intelligent embodied systems.
\newblock \emph{arXiv preprint arXiv:2503.06669}, 2025.

\bibitem[Cadene et~al.(2024)Cadene, Alibert, Soare, Gallouedec, Zouitine, Palma, Kooijmans, Aractingi, Shukor, Aubakirova, et~al.]{lerobot}
Remi Cadene, Simon Alibert, Alexander Soare, Quentin Gallouedec, Adil Zouitine, Steven Palma, Pepijn Kooijmans, Michel Aractingi, Mustafa Shukor, Dana Aubakirova, et~al.
\newblock {LeRobot}: State-of-the-art machine learning for real-world robotics in {PyTorch}.
\newblock \url{https://github.com/huggingface/lerobot}, 2024.

\bibitem[Carion et~al.(2026)Carion, Gustafson, Hu, Debnath, Hu, Coll-Vinent, Ryali, Alwala, Khedr, Huang, Lei, Ma, Guo, Kalla, Marks, Greer, Wang, Sun, R{\"a}dle, Afouras, Mavroudi, Xu, Wu, Zhou, Momeni, Hazra, Ding, Vaze, Porcher, Li, Li, Kamath, Cheng, Dollar, Ravi, Saenko, Zhang, and Feichtenhofer]{carion2026sam}
Nicolas Carion, Laura Gustafson, Yuan-Ting Hu, Shoubhik Debnath, Ronghang Hu, Didac~Suris Coll-Vinent, Chaitanya Ryali, Kalyan~Vasudev Alwala, Haitham Khedr, Andrew Huang, Jie Lei, Tengyu Ma, Baishan Guo, Arpit Kalla, Markus Marks, Joseph Greer, Meng Wang, Peize Sun, Roman R{\"a}dle, Triantafyllos Afouras, Effrosyni Mavroudi, Katherine Xu, Tsung-Han Wu, Yu~Zhou, Liliane Momeni, Rishi Hazra, Shuangrui Ding, Sagar Vaze, Francois Porcher, Feng Li, Siyuan Li, Aishwarya Kamath, Ho~Kei Cheng, Piotr Dollar, Nikhila Ravi, Kate Saenko, Pengchuan Zhang, and Christoph Feichtenhofer.
\newblock {SAM} 3: Segment anything with concepts.
\newblock In \emph{The Fourteenth International Conference on Learning Representations}, 2026.

\bibitem[Chen et~al.(2024)Chen, Chen, Zhang, Wang, Liu, Zhou, Zhang, Wan, Zhou, and Sun]{mllmjudge2024}
Dongping Chen, Ruoxi Chen, Shilin Zhang, Yaochen Wang, Yinuo Liu, Huichi Zhou, Qihui Zhang, Yao Wan, Pan Zhou, and Lichao Sun.
\newblock {MLLM-as-a-Judge}: Assessing multimodal {LLM-as-a-Judge} with vision-language benchmark.
\newblock In \emph{International Conference on Machine Learning}, volume 235 of \emph{Proceedings of Machine Learning Research}, pp.\  6562--6595. PMLR, 2024.

\bibitem[Chen et~al.(2022)Chen, Yi, Li, Ding, Torralba, Tenenbaum, and Gan]{comphy2022}
Zhenfang Chen, Kexin Yi, Yunzhu Li, Mingyu Ding, Antonio Torralba, Joshua~B. Tenenbaum, and Chuang Gan.
\newblock {ComPhy}: Compositional physical reasoning of objects and events from videos.
\newblock In \emph{International Conference on Learning Representations}, 2022.

\bibitem[Chow et~al.(2025)Chow, Mao, Li, Seita, Guizilini, and Wang]{physbench2025}
Wei Chow, Jiageng Mao, Boyi Li, Daniel Seita, Vitor~Campagnolo Guizilini, and Yue Wang.
\newblock {PhysBench}: Benchmarking and enhancing vision-language models for physical world understanding.
\newblock In \emph{The Thirteenth International Conference on Learning Representations}, 2025.

\bibitem[Deng et~al.(2026)Deng, Pan, Zhang, Li, Hu, Ding, Zou, Zeng, and Zhou]{RBench_deng2026rethinkingvideogenerationmodel}
Yufan Deng, Zilin Pan, Hongyu Zhang, Xiaojie Li, Ruoqing Hu, Yufei Ding, Yiming Zou, Yan Zeng, and Daquan Zhou.
\newblock Rethinking video generation model for the embodied world, 2026.
\newblock URL \url{https://arxiv.org/abs/2601.15282}.

\bibitem[Du et~al.(2023)Du, Yang, Dai, Dai, Nachum, Tenenbaum, Schuurmans, and Abbeel]{du2023unipi}
Yilun Du, Mengjiao Yang, Bo~Dai, Hanjun Dai, Ofir Nachum, Joshua~B. Tenenbaum, Dale Schuurmans, and Pieter Abbeel.
\newblock Learning universal policies via text-guided video generation.
\newblock In \emph{Advances in Neural Information Processing Systems}, volume~36, 2023.

\bibitem[Fang et~al.(2024)Fang, Fang, Tang, Liu, Wang, Wang, Zhu, and Lu]{rh20t}
Hao-Shu Fang, Hongjie Fang, Zhenyu Tang, Jirong Liu, Chenxi Wang, Junbo Wang, Haoyi Zhu, and Cewu Lu.
\newblock {RH20T}: A comprehensive robotic dataset for learning diverse skills in one-shot.
\newblock In \emph{IEEE International Conference on Robotics and Automation (ICRA)}, pp.\  653--660, 2024.

\bibitem[Farneb{\"a}ck(2003)]{farneback2003two}
Gunnar Farneb{\"a}ck.
\newblock Two-frame motion estimation based on polynomial expansion.
\newblock In \emph{Scandinavian Conference on Image Analysis (SCIA)}, volume 2749 of \emph{Lecture Notes in Computer Science}, pp.\  363--370. Springer, 2003.

\bibitem[Fu et~al.(2025)Fu, Zhao, and Finn]{mobilealoha}
Zipeng Fu, Tony~Z. Zhao, and Chelsea Finn.
\newblock Mobile {ALOHA}: Learning bimanual mobile manipulation using low-cost whole-body teleoperation.
\newblock In \emph{Proceedings of The 8th Conference on Robot Learning (CoRL)}, volume 270 of \emph{Proceedings of Machine Learning Research}, pp.\  4066--4083. PMLR, 2025.

\bibitem[Gao et~al.(2026)Gao, Liang, Zheng, Malik, Ye, Yu, Tseng, Dong, Mo, Lin, et~al.]{dreamdojo2026}
Shenyuan Gao, William Liang, Kaiyuan Zheng, Ayaan Malik, Seonghyeon Ye, Sihyun Yu, Wei-Cheng Tseng, Yuzhu Dong, Kaichun Mo, Chen-Hsuan Lin, et~al.
\newblock {DreamDojo}: A generalist robot world model from large-scale human videos.
\newblock In \emph{International Conference on Machine Learning (ICML)}, 2026.

\bibitem[Gillman et~al.(2025)Gillman, Herrmann, Freeman, Aggarwal, Luo, Sun, and Sun]{forceprompting2025}
Nate Gillman, Charles Herrmann, Michael Freeman, Daksh Aggarwal, Evan Luo, Deqing Sun, and Chen Sun.
\newblock Force prompting: Video generation models can learn and generalize physics-based control signals.
\newblock In \emph{Advances in Neural Information Processing Systems}, volume~38, 2025.

\bibitem[Gu et~al.(2026)Gu, Liu, Zeng, Nagarajan, Zhu, Hong, Fan, Yan, Zhou, Liu, and Wang]{gu2026phyworldbenchcomprehensiveevaluationphysical}
Jing Gu, Xian Liu, Yu~Zeng, Ashwin Nagarajan, Fangrui Zhu, Daniel Hong, Yue Fan, Qianqi Yan, Kaiwen Zhou, Ming-Yu Liu, and Xin~Eric Wang.
\newblock ``{PhyWorldBench}'': A comprehensive evaluation of physical realism in text-to-video models, 2026.
\newblock URL \url{https://arxiv.org/abs/2507.13428}.

\bibitem[Hong et~al.(2025)Hong, Cheng, Yang, Wang, Wang, Gu, Huang, Dong, and Tang]{hong2025motionbench}
Wenyi Hong, Yean Cheng, Zhuoyi Yang, Weihan Wang, Lefan Wang, Xiaotao Gu, Shiyu Huang, Yuxiao Dong, and Jie Tang.
\newblock {MotionBench}: Benchmarking and improving fine-grained video motion understanding for vision language models.
\newblock In \emph{Proceedings of the Computer Vision and Pattern Recognition Conference}, pp.\  8450--8460, 2025.

\bibitem[Huang et~al.(2024)Huang, He, Yu, Zhang, Si, et~al.]{huang2024vbench}
Ziqi Huang, Yinan He, Jiashuo Yu, Fan Zhang, Chenyang Si, et~al.
\newblock {VBench}: Comprehensive benchmark suite for video generative models.
\newblock In \emph{CVPR}, 2024.

\bibitem[Internò et~al.(2026)Internò, Pondaven, Issa, Pizzati, Pinto, Olhofer, Laptev, Torr, Simoncelli, Hammer, and Klindt]{internò2026geophysgeometryphysicalplausibility}
Christian Internò, Alexander Pondaven, Habon Issa, Fabio Pizzati, Francesco Pinto, Markus Olhofer, Ivan Laptev, Philip Torr, Eero~P. Simoncelli, Barbara Hammer, and David Klindt.
\newblock {GEOPHYS}: The geometry of physical plausibility, 2026.
\newblock URL \url{https://arxiv.org/abs/2606.20707}.

\bibitem[Jang et~al.(2025)Jang, Ye, Lin, Xiang, Bjorck, Fang, Hu, Huang, Kundalia, Lin, et~al.]{dreamgen2025}
Joel Jang, Seonghyeon Ye, Zongyu Lin, Jiannan Xiang, Johan Bjorck, Yu~Fang, Fengyuan Hu, Spencer Huang, Kaushil Kundalia, Yen-Chen Lin, et~al.
\newblock {DreamGen}: Unlocking generalization in robot learning through video world models.
\newblock In \emph{Conference on Robot Learning}, volume 305 of \emph{Proceedings of Machine Learning Research}, pp.\  5170--5194. PMLR, 2025.

\bibitem[Jiang et~al.(2026)Jiang, Chen, Xu, Liu, Wang, Shen, Lu, Huang, Wang, Xie, and Wu]{jiang2026robowmbenchbenchmarkevaluatingworld}
Feng Jiang, Yang Chen, Kyle Xu, Yuchen Liu, Haifeng Wang, Zhenhao Shen, Jasper Lu, Shengze Huang, Yuanfei Wang, Chen Xie, and Ruihai Wu.
\newblock {RoboWM-Bench}: A benchmark for evaluating world models in robotic manipulation, 2026.
\newblock URL \url{https://arxiv.org/abs/2604.19092}.

\bibitem[Kang et~al.(2025)Kang, Yue, Lu, Lin, Zhao, Wang, Huang, and Feng]{kang2025howfar}
Bingyi Kang, Yang Yue, Rui Lu, Zhijie Lin, Yang Zhao, Kaixin Wang, Gao Huang, and Jiashi Feng.
\newblock How far is video generation from world model: A physical law perspective.
\newblock In \emph{Proceedings of the 42nd International Conference on Machine Learning}, volume 267 of \emph{Proceedings of Machine Learning Research}, pp.\  28991--29017. PMLR, 2025.
\newblock URL \url{https://proceedings.mlr.press/v267/kang25g.html}.

\bibitem[Le et~al.(2025)Le, Zhu, Kalogeiton, and Samaras]{le2025gravity}
Minh-Quan Le, Yuanzhi Zhu, Vicky Kalogeiton, and Dimitris Samaras.
\newblock What about gravity in video generation? post-training {Newton's} laws with verifiable rewards.
\newblock \emph{arXiv preprint arXiv:2512.00425}, 2025.

\bibitem[Li et~al.(2025)Li, Fang, Chen, Yang, Cao, Wong, Luo, Wang, Yin, Gonzalez, Stoica, Han, and Lu]{li2025worldmodelbenchjudgingvideogeneration}
Dacheng Li, Yunhao Fang, Yukang Chen, Shuo Yang, Shiyi Cao, Justin Wong, Michael Luo, Xiaolong Wang, Hongxu Yin, Joseph~E. Gonzalez, Ion Stoica, Song Han, and Yao Lu.
\newblock {WorldModelBench}: Judging video generation models as world models, 2025.
\newblock URL \url{https://arxiv.org/abs/2502.20694}.

\bibitem[Li et~al.(2026)Li, Wen, Song, Zhu, Hao, Chen, Chen, Wu, Chu, and Wang]{phyparam2026}
Yanxun Li, Hao Wen, Bingze Song, Jiashu Zhu, Aiming Hao, Chubin Chen, Jintao Chen, Jiahong Wu, Xiangxiang Chu, and Miao Wang.
\newblock Learning explicit physical parameter control and benchmarking for video generation.
\newblock \emph{arXiv preprint arXiv:2607.18924}, 2026.

\bibitem[Liu et~al.(2024)Liu, Ren, Gupta, and Wang]{physgen2024}
Shaowei Liu, Zhongzheng Ren, Saurabh Gupta, and Shenlong Wang.
\newblock {PhysGen}: Rigid-body physics-grounded image-to-video generation.
\newblock In \emph{European Conference on Computer Vision}, volume 15140 of \emph{Lecture Notes in Computer Science}, pp.\  360--378. Springer, 2024.

\bibitem[Lucas \& Kanade(1981)Lucas and Kanade]{lucas1981iterative}
Bruce~D. Lucas and Takeo Kanade.
\newblock An iterative image registration technique with an application to stereo vision.
\newblock In \emph{Proceedings of the 7th International Joint Conference on Artificial Intelligence (IJCAI)}, pp.\  674--679, 1981.

\bibitem[Meng et~al.(2025)Meng, Liao, Tan, Shao, Lu, Zhang, Cheng, Li, Qiao, and Luo]{meng2024phygenbench}
Fanqing Meng, Jiaqi Liao, Xinyu Tan, Wenqi Shao, Quanfeng Lu, Kaipeng Zhang, Yu~Cheng, Dianqi Li, Yu~Qiao, and Ping Luo.
\newblock Towards world simulator: Crafting physical commonsense-based benchmark for video generation.
\newblock In \emph{ICML}, 2025.

\bibitem[Motamed et~al.(2026)Motamed, Culp, Swersky, Jaini, and Geirhos]{motamed2026physicsiq}
Saman Motamed, Laura Culp, Kevin Swersky, Priyank Jaini, and Robert Geirhos.
\newblock Do generative video models understand physical principles?
\newblock In \emph{WACV}, pp.\  948--958, 2026.

\bibitem[Nguyen et~al.(2026)Nguyen, Diep, Nguyen, Le, Thien, Tran, Le, Duong, Sonntag, Le, et~al.]{nguyen2026robogaze}
Minh-Loi Nguyen, Nghiem~Tuong Diep, Hung~Khang Nguyen, Minh Le, Doanh~Le Thien, Hoang~H Tran, Dung~D Le, Vu~N Duong, Daniel Sonntag, An~Thai Le, et~al.
\newblock {RoboGaze}: Evaluating robot world models via structured vision-language analysis.
\newblock \emph{arXiv preprint arXiv:2606.28385}, 2026.

\bibitem[{NVIDIA}(2025{\natexlab{a}})]{cosmoswfm2025}
{NVIDIA}.
\newblock {Cosmos} world foundation model platform for physical {AI}.
\newblock \emph{arXiv preprint arXiv:2501.03575}, 2025{\natexlab{a}}.

\bibitem[{NVIDIA}(2025{\natexlab{b}})]{nvidia2025cosmosreason}
{NVIDIA}.
\newblock {Cosmos-Reason1}: From physical common sense to embodied reasoning.
\newblock \emph{arXiv preprint arXiv:2503.15558}, 2025{\natexlab{b}}.

\bibitem[{Open X-Embodiment Collaboration} et~al.(2024){Open X-Embodiment Collaboration}, O'Neill, Rehman, Maddukuri, Gupta, Padalkar, Lee, Pooley, Gupta, Mandlekar, et~al.]{oxe}
{Open X-Embodiment Collaboration}, Abby O'Neill, Abdul Rehman, Abhiram Maddukuri, Abhishek Gupta, Abhishek Padalkar, Abraham Lee, Acorn Pooley, Agrim Gupta, Ajay Mandlekar, et~al.
\newblock Open {X-Embodiment}: Robotic learning datasets and {RT-X} models.
\newblock In \emph{IEEE International Conference on Robotics and Automation (ICRA)}, pp.\  6892--6903, 2024.
\newblock \doi{10.1109/ICRA57147.2024.10611477}.

\bibitem[{OpenAI}(2024)]{openai2024gpt4o}
{OpenAI}.
\newblock {GPT-4o} system card, 2024.
\newblock URL \url{https://arxiv.org/abs/2410.21276}.

\bibitem[Oquab et~al.(2024)Oquab, Darcet, Moutakanni, Vo, Szafraniec, Khalidov, Fernandez, Haziza, Massa, El-Nouby, et~al.]{oquab2024dinov2}
Maxime Oquab, Timoth{\'e}e Darcet, Th{\'e}o Moutakanni, Huy~V. Vo, Marc Szafraniec, Vasil Khalidov, Pierre Fernandez, Daniel Haziza, Francisco Massa, Alaaeldin El-Nouby, et~al.
\newblock {DINOv2}: Learning robust visual features without supervision.
\newblock \emph{Transactions on Machine Learning Research}, 2024.

\bibitem[Padalkar et~al.(2023)Padalkar, Quere, Steinmetz, Raffin, Nieuwenhuisen, Silv{\'e}rio, and Stulp]{dlrsara}
Abhishek Padalkar, Gabriel Quere, Franz Steinmetz, Antonin Raffin, Matthias Nieuwenhuisen, Jo{\~a}o Silv{\'e}rio, and Freek Stulp.
\newblock Guiding reinforcement learning with shared control templates.
\newblock In \emph{IEEE International Conference on Robotics and Automation (ICRA)}, 2023.
\newblock \doi{10.1109/ICRA48891.2023.10161058}.

\bibitem[Pothiraj et~al.(2026)Pothiraj, Cho, Zhang, Stengel-Eskin, and Bansal]{pothiraj2026physicsquestionscenegraph}
Atin Pothiraj, Jaemin Cho, Yue Zhang, Elias Stengel-Eskin, and Mohit Bansal.
\newblock Physics question scene graph: Fine-grained evaluation of physical plausibility in text-to-video generation, 2026.
\newblock URL \url{https://arxiv.org/abs/2606.25306}.

\bibitem[Puyin et~al.(2026)Puyin, Xiang, Mao, Wei, Chen, Masood, Fei-Fei, and Adeli]{puyin2026quantiphy}
Li~Puyin, Tiange Xiang, Ella Mao, Shirley Wei, Xinye Chen, Adnan Masood, Li~Fei-Fei, and Ehsan Adeli.
\newblock {QuantiPhy}: A quantitative benchmark evaluating physical reasoning abilities of vision-language models.
\newblock In \emph{Proceedings of the IEEE/CVF Conference on Computer Vision and Pattern Recognition}, pp.\  33174--33184, 2026.

\bibitem[Qin et~al.(2025)Qin, Shi, Yu, Wang, Zhou, Li, Yin, Liu, Sheng, Shao, Bai, and Zhang]{qin2025worldsimbench}
Yiran Qin, Zhelun Shi, Jiwen Yu, Xijun Wang, Enshen Zhou, Lijun Li, Zhenfei Yin, Xihui Liu, Lu~Sheng, Jing Shao, Lei Bai, and Ruimao Zhang.
\newblock {WorldSimBench}: Towards video generation models as world simulators.
\newblock In \emph{Forty-second International Conference on Machine Learning}, 2025.
\newblock URL \url{https://openreview.net/forum?id=j9pVnmulQm}.

\bibitem[Quevedo et~al.(2025)Quevedo, Sharma, Sun, Suryavanshi, Liang, and Yang]{quevedo2025worldgym}
Julian Quevedo, Ansh~Kumar Sharma, Yixiang Sun, Varad Suryavanshi, Percy Liang, and Sherry Yang.
\newblock {WorldGym}: World model as an environment for policy evaluation.
\newblock \emph{arXiv preprint arXiv:2506.00613}, 2025.

\bibitem[Sun et~al.(2025)Sun, Huang, Liu, Wu, Xu, Li, and Liu]{sun2025t2vcompbenchcomprehensivebenchmarkcompositional}
Kaiyue Sun, Kaiyi Huang, Xian Liu, Yue Wu, Zihan Xu, Zhenguo Li, and Xihui Liu.
\newblock {T2V-CompBench}: A comprehensive benchmark for compositional text-to-video generation, 2025.
\newblock URL \url{https://arxiv.org/abs/2407.14505}.

\bibitem[{Team Wan} et~al.(2025){Team Wan}, Wang, Ai, Wen, Mao, Xie, Chen, Yu, Zhao, Yang, et~al.]{wan2025wan}
{Team Wan}, Ang Wang, Baole Ai, Bin Wen, Chaojie Mao, Chen-Wei Xie, Di~Chen, Feiwu Yu, Haiming Zhao, Jianxiao Yang, et~al.
\newblock Wan: Open and advanced large-scale video generative models.
\newblock \emph{arXiv preprint arXiv:2503.20314}, 2025.

\bibitem[Teng et~al.(2025)Teng, Jia, Sun, Li, Li, Tang, Han, Zhang, Zhang, Luo, et~al.]{teng2025magi}
Hansi Teng, Hongyu Jia, Lei Sun, Lingzhi Li, Maolin Li, Mingqiu Tang, Shuai Han, Tianning Zhang, WQ~Zhang, Weifeng Luo, et~al.
\newblock {MAGI-1}: Autoregressive video generation at scale.
\newblock \emph{arXiv preprint arXiv:2505.13211}, 2025.

\bibitem[Tung et~al.(2023)Tung, Ding, Chen, Bear, Gan, Tenenbaum, Yamins, Fan, and Smith]{tung2023physionevaluatingphysicalscene}
Hsiao-Yu Tung, Mingyu Ding, Zhenfang Chen, Daniel Bear, Chuang Gan, Joshua~B. Tenenbaum, Daniel~LK Yamins, Judith~E Fan, and Kevin~A. Smith.
\newblock Physion++: Evaluating physical scene understanding that requires online inference of different physical properties, 2023.
\newblock URL \url{https://arxiv.org/abs/2306.15668}.

\bibitem[{Unitree Robotics}(2025{\natexlab{a}})]{unitree_g1dex1}
{Unitree Robotics}.
\newblock {G1\_Dex1} datasets.
\newblock \url{https://huggingface.co/datasets/unitreerobotics/G1_Dex1_MountCamera_Dataset}, 2025{\natexlab{a}}.
\newblock Hugging Face dataset repositories unitreerobotics/G1\_Dex1\_*.

\bibitem[{Unitree Robotics}(2025{\natexlab{b}})]{unitree_g1dex3}
{Unitree Robotics}.
\newblock {G1\_Dex3} datasets.
\newblock \url{https://huggingface.co/datasets/unitreerobotics/G1_Dex3_BlockStacking_Dataset}, 2025{\natexlab{b}}.
\newblock Hugging Face dataset repositories unitreerobotics/G1\_Dex3\_*.

\bibitem[{Unitree Robotics}(2025{\natexlab{c}})]{unitree_z1dual}
{Unitree Robotics}.
\newblock {Z1\_Dual\_Dex1\_PourCoffee\_Dataset}.
\newblock \url{https://huggingface.co/datasets/unitreerobotics/Z1_Dual_Dex1_PourCoffee_Dataset}, 2025{\natexlab{c}}.

\bibitem[Unterthiner et~al.(2018)Unterthiner, Van~Steenkiste, Kurach, Marinier, Michalski, and Gelly]{unterthiner2018towards}
Thomas Unterthiner, Sjoerd Van~Steenkiste, Karol Kurach, Raphael Marinier, Marcin Michalski, and Sylvain Gelly.
\newblock Towards accurate generative models of video: A new metric \& challenges.
\newblock \emph{arXiv preprint arXiv:1812.01717}, 2018.

\bibitem[Wang et~al.(2025)Wang, Xu, Cheng, Diao, Zhou, Cao, Wang, Ge, and Huang]{wang2025grounded}
Haibo Wang, Zhiyang Xu, Yu~Cheng, Shizhe Diao, Yufan Zhou, Yixin Cao, Qifan Wang, Weifeng Ge, and Lifu Huang.
\newblock Grounded-{V}ideo{LLM}: Sharpening fine-grained temporal grounding in video large language models.
\newblock In Christos Christodoulopoulos, Tanmoy Chakraborty, Carolyn Rose, and Violet Peng (eds.), \emph{Findings of the Association for Computational Linguistics: EMNLP 2025}, pp.\  959--975, Suzhou, China, November 2025. Association for Computational Linguistics.
\newblock ISBN 979-8-89176-335-7.

\bibitem[Wu et~al.(2025{\natexlab{a}})Wu, Zou, Li, Huang, Yang, Tan, Peng, Wu, Xiong, Jiang, et~al.]{wu2025hunyuanvideo}
Bing Wu, Chang Zou, Changlin Li, Duojun Huang, Fang Yang, Hao Tan, Jack Peng, Jianbing Wu, Jiangfeng Xiong, Jie Jiang, et~al.
\newblock {HunyuanVideo 1.5} technical report.
\newblock \emph{arXiv preprint arXiv:2511.18870}, 2025{\natexlab{a}}.

\bibitem[Wu et~al.(2025{\natexlab{b}})Wu, Liu, Xie, Wang, Li, Yang, Li, Zhu, Wu, Liu, et~al.]{robocoin}
Shihan Wu, Xuecheng Liu, Shaoxuan Xie, Pengwei Wang, Xinghang Li, Bowen Yang, Zhe Li, Kai Zhu, Hongyu Wu, Yiheng Liu, et~al.
\newblock {RoboCOIN}: An open-sourced bimanual robotic data collection for integrated manipulation.
\newblock \emph{arXiv preprint arXiv:2511.17441}, 2025{\natexlab{b}}.

\bibitem[Wu et~al.(2026)Wu, Li, Jin, KV, Raj, Sinha, Chen, Du, and Manocha]{wu2026mass}
Xiyang Wu, Zongxia Li, Jihui Jin, Gouthaman KV, Vishnu Raj, Nilotpal Sinha, Jingxi Chen, Fan Du, and Dinesh Manocha.
\newblock {MASS}: Motion-aware spatial-temporal grounding for physics reasoning and comprehension in vision-language models.
\newblock In \emph{Proceedings of the IEEE/CVF Conference on Computer Vision and Pattern Recognition}, pp.\  9433--9443, 2026.

\bibitem[Yang et~al.(2024)Yang, Du, Ghasemipour, Tompson, Kaelbling, Schuurmans, and Abbeel]{unisim2024}
Sherry Yang, Yilun Du, Kamyar Ghasemipour, Jonathan Tompson, Leslie Kaelbling, Dale Schuurmans, and Pieter Abbeel.
\newblock Learning interactive real-world simulators.
\newblock In \emph{International Conference on Learning Representations}, 2024.

\bibitem[Yue et~al.(2025)Yue, Huang, Liao, Chen, Zhou, Chen, Yao, and Ren]{yue2025ewmbench}
Hu~Yue, Siyuan Huang, Yue Liao, Shengcong Chen, Pengfei Zhou, Liliang Chen, Maoqing Yao, and Guanghui Ren.
\newblock {EWMBench}: Evaluating scene, motion, and semantic quality in embodied world models.
\newblock \emph{arXiv preprint arXiv:2505.09694}, 2025.

\bibitem[Zhang et~al.(2025)Zhang, Cherniavskii, et~al.]{zhang2025morpheus}
Chenyu Zhang, Daniil Cherniavskii, et~al.
\newblock Morpheus: Benchmarking physical reasoning of video generative models with real physical experiments.
\newblock \emph{arXiv preprint arXiv:2504.02918}, 2025.

\bibitem[Zhang et~al.(2026)Zhang, Jing, Yu, Ding, Nie, Wang, Du, Zou, Wu, and Shuai]{zhang2026physionevalevaluatingphysicalrealism}
Qin Zhang, Peiyu Jing, Hong-Xing Yu, Fangqiang Ding, Fan Nie, Weimin Wang, Yilun Du, James Zou, Jiajun Wu, and Bing Shuai.
\newblock {Physion-Eval}: Evaluating physical realism in generated video via human reasoning, 2026.
\newblock URL \url{https://arxiv.org/abs/2603.19607}.

\bibitem[Zhao et~al.(2023)Zhao, Kumar, Levine, and Finn]{aloha}
Tony~Z. Zhao, Vikash Kumar, Sergey Levine, and Chelsea Finn.
\newblock Learning fine-grained bimanual manipulation with low-cost hardware.
\newblock In \emph{Proceedings of Robotics: Science and Systems (RSS)}, 2023.
\newblock \doi{10.15607/RSS.2023.XIX.016}.

\bibitem[Zhao et~al.(2026)Zhao, Jing, Liu, Mao, Jha, Yang, Xue, Zakharov, Guizilini, and Wang]{humanoideveryday}
Zhenyu Zhao, Hongyi Jing, Xiawei Liu, Jiageng Mao, Abha Jha, Hanwen Yang, Rong Xue, Sergey Zakharov, Vitor Guizilini, and Yue Wang.
\newblock {Humanoid Everyday}: A comprehensive robotic dataset for open-world humanoid manipulation.
\newblock In \emph{IEEE International Conference on Robotics and Automation (ICRA)}, 2026.

\bibitem[Zheng et~al.(2025)Zheng, Huang, Liu, Zou, He, Zhang, Gu, Zhang, He, Zheng, et~al.]{zheng2025vbench}
Dian Zheng, Ziqi Huang, Hongbo Liu, Kai Zou, Yinan He, Fan Zhang, Lulu Gu, Yuanhan Zhang, Jingwen He, Wei-Shi Zheng, et~al.
\newblock {VBench-2.0}: Advancing video generation benchmark suite for intrinsic faithfulness.
\newblock \emph{arXiv preprint arXiv:2503.21755}, 2025.

\end{thebibliography}
\bibliographystyle{iclr2027_conference}

\clearpage
\appendix
\section{Dataset Details}
\label{app:dataset-details}
\subsection{Source Datasets and Licenses}
\label{app:sources}
 
The 80 starting frames come from demonstrations in the 11 public robot datasets listed in Table~\ref{tab:sources}. For each scenario we use one frame from one demonstration, taken before the manipulation begins; \texttt{manifest.csv} in the release gives the source, episode, and frame index of every frame. Most sources are LeRobot-format releases \citep{lerobot}. Licenses are those stated by each source, and every released file keeps its source license. Our own prompts and annotations are available on Hugging Face. 
 
\begin{table}[ht]
\centering
\scriptsize
\setlength{\tabcolsep}{4pt}
\def\UrlBreaks{\do\/\do\-\do\_\do\.}%
\caption{Public datasets that supply the starting frames. Source is the Hugging Face repository we used; $^\dagger$ marks an unofficial copy, with the official release in parentheses.}
\label{tab:sources}
\begin{tabular}{@{}>{\raggedright\arraybackslash}p{0.21\linewidth}>{\raggedright\arraybackslash}p{0.25\linewidth}>{\raggedright\arraybackslash}p{0.21\linewidth}>{\raggedright\arraybackslash}p{\dimexpr0.33\linewidth-6\tabcolsep\relax}@{}}
\toprule
Dataset & Robot (type) & License & Source \\
\midrule
Humanoid Everyday \citep{humanoideveryday} & Unitree G1, H1 (humanoid) & Apache-2.0 & \path{USC-GVL/humanoid-everyday} \\
RoboCOIN \citep{robocoin} & AlphaBot 2, Galbot G1, Unitree G1edu (dual-arm, humanoid) & Apache-2.0, with use terms & \path{RoboCOIN/*} \\
Unitree G1 Dex1 \citep{unitree_g1dex1} & Unitree G1 with Dex1 grippers (humanoid) & Apache-2.0; none stated for UniBot-V1 challenge sets & \path{unitreerobotics/G1_Dex1_*} \\
Unitree G1 Dex3 \citep{unitree_g1dex3} & Unitree G1 with Dex3 hands (humanoid) & Apache-2.0 & \path{unitreerobotics/G1_Dex3_*} \\
Unitree Z1 dual-arm \citep{unitree_z1dual} & Two Unitree Z1 arms (dual-arm) & Apache-2.0 & \path{unitreerobotics/Z1_Dual_Dex1_PourCoffee_Dataset} \\
GR00T-Teleop-GR1 \citep{dreamdojo2026} & Fourier GR-1 (humanoid) & CC BY-NC 4.0 & \path{nvidia/PhysicalAI-Robotics-GR00T-Teleop-GR1} \\
AgiBot World Alpha \citep{agibotworld} & AgiBot G1 (dual-arm, mobile) & CC BY-NC-SA 4.0 & \path{weijian-sun/agibotworld-lerobot}$^\dagger$ (\path{agibot-world/AgiBotWorld-Alpha}) \\
Mobile ALOHA \citep{mobilealoha} & Mobile ALOHA (dual-arm, mobile) & MIT or Apache-2.0, by repository & \path{lerobot/aloha_mobile_*} \\
ALOHA \citep{aloha} & Static ALOHA (dual-arm) & MIT or Apache-2.0, by repository & \path{lerobot/aloha_static_*} \\
RH20T \citep{rh20t} & Flexiv arm (single-arm) & CC BY-SA 4.0 or CC BY-NC 4.0, by scene & \path{robot-lev/rh20t_cfg1}$^\dagger$ (\path{rh20t.github.io}) \\
DLR SARA Pour \citep{dlrsara} & DLR SARA (single-arm) & CC BY 4.0 & \path{lerobot/dlr_sara_pour}, from Open X-Embodiment \citep{oxe} \\
\bottomrule
\end{tabular}
\end{table}

\subsection{Image and Prompt Preparation}
We took each conditioning frame from before the target manipulation and standardized the supplied frames to \(1280\times720\) pixels. GPT-4o~\citep{openai2024gpt4o} received each frame and its logged task under a fixed two-field JSON instruction. It produced a description of the visible scene and a one-sentence description of the complete manipulation without specifying its outcome. We reviewed these descriptions and assembled the generation prompt from the \texttt{Scene} and \texttt{Action} fields. For unobservable variants, we retained the same frame and action while rewriting the scene description to state the hidden property. Expected outcomes and failure signatures were used for evaluation and excluded from the generation prompt.

\subsection{Video Generation}
\label{app:generation}

Each model receives the same \(1280\times720\) conditioning frame and text prompt, and we generate one video per prompt and model with the settings in Table~\ref{tab:generation}.

\begin{table}[ht]
\centering
\small
\setlength{\tabcolsep}{4pt}
\caption{Video generation settings. All clips are image-to-video, conditioned on a
start frame from the source clip plus a text prompt, and last about 5\,s.}
\label{tab:generation}
\begin{tabular}{@{}lllllll@{}}
\toprule
Model & Checkpoint & Resolution$^{a}$ & Frames (fps) & Steps & Guidance & Seed \\
\midrule
Wan 2.2          & Wan2.2-TI2V-5B            & 960$\times$704 & 121 (24) & 40         & 5.0        & 0 \\
Cosmos 3 Nano    & Cosmos3-Nano              & 960$\times$720 & 121 (24) & 35         & 6.0        & 0 \\
HunyuanVideo 1.5 & HunyuanVideo-1.5-480p-I2V & 720$\times$544 & 121 (24) & 50         & 6.0        & 0 \\
MAGI 4.5B Distill & MAGI-1 4.5B distill      & 960$\times$720 & 120 (24) & 16$^{b}$   & --$^{c}$   & 0 \\
\bottomrule
\end{tabular}
\par\smallskip
\parbox{\linewidth}{\footnotesize
$^{a}$Width follows the source clip's aspect ratio; the size shown is for the 4:3 majority.
$^{b}$MAGI-1 generates autoregressively in 24-frame chunks, with 16 denoising steps per chunk.
$^{c}$Distilled checkpoint with no classifier-free guidance.
Only Wan uses a negative prompt (its default); the other models use none.
HunyuanVideo uses flow shift 5.0.
}
\end{table}

\subsection{Taxonomy and Curation Notes}
Collision and Friction scenarios were comparatively easy to source, reflecting the prevalence of pick-and-place and surface-manipulation footage. Fluid and Deformation scenarios were harder to find, while Gravity scenarios that isolated free fall were rare outside incidental object drops. These observations describe our curation process only; establishing the distribution of public robotics video would require a systematic survey of available sources.

\paragraph{Annotation procedure.} Scenarios were curated from public robotics video sources rather than authored domain-first, and each scenario was subsequently annotated with its governing physical domain by manual review of its action and outcome. This annotation was performed independently of the taxonomy's construction to avoid circularity: the seven domains were fixed in advance from the tool's seven specialists (Section~\ref{sec:pipeline}), and each scenario was assigned to the domain its physics most directly tests, rather than domains being defined post hoc to fit whatever scenarios happened to be available.

\paragraph{Interaction with observability.} The observable/unobservable axis introduced in \S\ref{sec:observability} is orthogonal to domain: a scenario's domain label is fixed by the physical law under test, while its observability condition determines whether the property governing that law is visually apparent or text-specified. Both matched members of an observable/unobservable pair therefore carry the same domain label.

\section{Human Annotation}
\label{app:annotation-details}

\subsection{Annotation Instrument}

\begin{figure}[t]
\centering
\includegraphics[width=\linewidth]{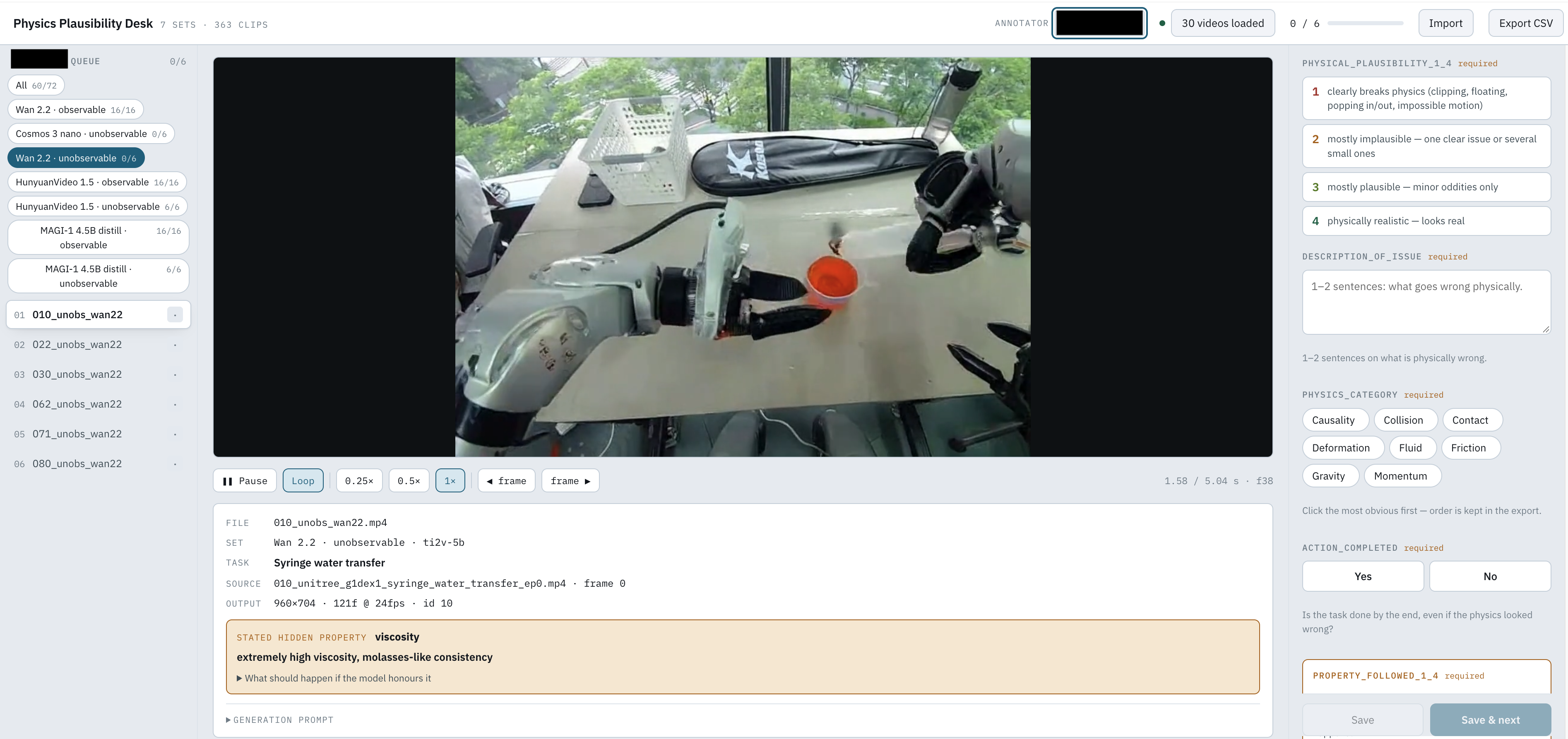}
\caption{The human-annotation interface. Annotators assess physical plausibility and action completion for every generated clip, identify visible violations when present, and evaluate property adherence for unobservable scenarios.}
\label{fig:annotation-interface}
\end{figure}

The annotation interface presents one generated clip at a time together with the fields summarized in Table~\ref{tab:annotation-fields}. Physical plausibility and action completion are collected for every clip. Hidden-property adherence is collected only for unobservable scenarios, for which the form includes a property-specific question. When a clip receives a plausibility score below 4, annotators must also describe the issue and select at least one violation category. Figure~\ref{fig:annotation-interface} shows the interface presented to annotators.

\begin{table}[t]
\centering
\small
\setlength{\tabcolsep}{4pt}
\renewcommand{\arraystretch}{1.1}
\caption{Fields in the human-annotation form. Monospace text gives each field's identifier. $P$: physical plausibility; $A$: action completion; $H$: hidden-property adherence.}
\label{tab:annotation-fields}
\begin{tabular}{@{}>{\raggedright\arraybackslash}p{0.22\linewidth}>{\raggedright\arraybackslash}p{0.12\linewidth}>{\raggedright\arraybackslash}p{0.13\linewidth}>{\raggedright\arraybackslash}p{\dimexpr0.53\linewidth-6\tabcolsep\relax}@{}}
\toprule
Field & Response & Collected for & Instructions \\
\midrule
Physical plausibility ($P$)\newline{\scriptsize\texttt{physical\_plausibility}}
  & 1--4
  & All clips
  & \textbf{1}: clearly breaks physics, including clipping, floating, objects appearing or disappearing, or impossible motion.\newline
    \textbf{2}: mostly implausible, with one clear issue or several smaller issues.\newline
    \textbf{3}: mostly plausible, with minor oddities only.\newline
    \textbf{4}: physically realistic. \\
\addlinespace
Issue description\newline{\scriptsize\texttt{description\_of\_issue}}
  & Free text, 1--2 sentences
  & Clips with $P<4$
  & Briefly describe what is physically incorrect. Leave blank when $P=4$. \\
\addlinespace
Violation labels\newline{\scriptsize\texttt{physics\_category}}
  & Multi-select, 8 labels
  & Clips with $P<4$
  & Select at least one applicable violation category, choosing the most obvious issue first. Leave blank when $P=4$. \\
\addlinespace
Action completion ($A$)\newline{\scriptsize\texttt{action\_completed}}
  & Yes/no
  & All clips
  & Judge whether the requested task is completed by the end of the clip, even if the process is physically unrealistic. \\
\addlinespace
Hidden-property adherence ($H$)\newline{\scriptsize\texttt{property\_followed}}
  & 1--4
  & Unobservable clips
  & \textbf{1}: property completely ignored; the outcome resembles a generic response.\newline
    \textbf{2}: mostly ignored, with only a faint or inconsistent indication.\newline
    \textbf{3}: mostly followed, but with incorrect magnitude, timing, or degree.\newline
    \textbf{4}: fully and correctly reflected. \\
\bottomrule
\end{tabular}
\end{table}

\begin{figure}[t]
  \centering
  \begin{minipage}[b]{0.32\textwidth}
    \centering
    \includegraphics[width=\linewidth]{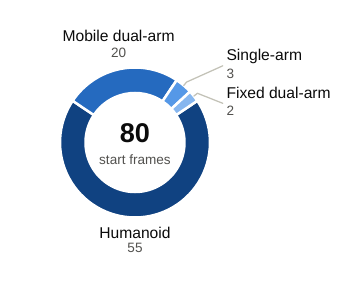}
    \subcaption{Robot type}\label{fig:ds-robots}
  \end{minipage}\hfill
  \begin{minipage}[b]{0.32\textwidth}
    \centering
    \includegraphics[width=\linewidth]{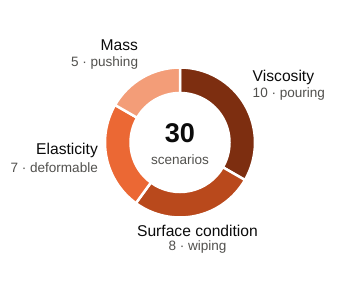}
    \subcaption{Hidden property}\label{fig:ds-hidden}
  \end{minipage}\hfill
  \begin{minipage}[b]{0.32\textwidth}
    \centering
    \includegraphics[width=\linewidth]{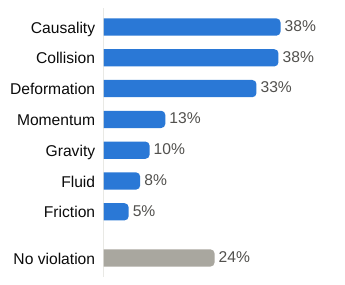}
    \subcaption{Violation label}\label{fig:ds-violations}
  \end{minipage}
  \caption{Dataset composition.
    (a)~Robot type in the 80 starting frames. Mobile dual-arm robots are wheeled platforms
    (AlphaBot~2, Galbot~G1, AgiBot~G1, and Mobile ALOHA).
    (b)~Property stated in each unobservable scenario and the task it is paired with.
    (c)~Share of the 439 videos with each violation label. As in Appendix~\ref{app:pipeline-details}, Contact is
    merged into Collision. A video can have several labels (66\% of videos with a violation have
    more than one), so the violation bars sum to over 100\%.}
  \label{fig:dataset-stats}
\end{figure}
\paragraph{Interpretation of the scores.}
The physical-plausibility score $P\in\{1,2,3,4\}$ assesses only the visible consistency of contact, motion, deformation, and object persistence. It is a human judgment rather than a measurement against a physical simulator. Task incompletion alone does not imply physical implausibility.

The action-completion score $A\in\{0,1\}$ assesses the final task outcome rather than the physical correctness of the process. A clip may therefore receive $A=1$ while exhibiting severe physical violations, or $A=0$ despite otherwise plausible motion.

The hidden-property score $H\in\{1,2,3,4\}$ measures whether the generated behavior demonstrates the property stated in the prompt. It is not collected for observable scenarios, and missing values in those scenarios are not interpreted as zero or as failures. Importantly, a low adherence score does not distinguish between two cases: the video may visibly contradict the stated property, or it may fail to perform an interaction that would reveal the property. Because the instrument does not provide a separate ``insufficient evidence'' response, we retain this ambiguity when interpreting $H\leq2$.

\paragraph{Aggregation.}
We report rating distributions, descriptive means of the ordinal scores, and action-completion proportions. For thresholded summaries, $P\geq3$ denotes mostly plausible or physically realistic behavior, while $H\geq3$ denotes that the stated property is mostly or fully followed. We additionally examine the joint distribution of $P$ and $H$, including the proportion of apparently plausible clips ($P\geq3$) that do not follow the hidden property ($H\leq2$). The three dimensions are reported separately rather than combined into a single quality score.

\subsection{Violation-Label Codebook}
\label{app:codebook}

The \texttt{physics\_category} field contains the eight labels defined in Table~\ref{tab:violation-labels}. It is shown for every clip and requires at least one selection when $P<4$. Annotators are instructed to select the most obvious violation first. Although click order is preserved in the exported data, our frequency analyses treat the selected labels as an unordered set; order is not interpreted as a calibrated ranking of severity.

\begin{table}[ht]
\centering
\small
\setlength{\tabcolsep}{5pt}
\renewcommand{\arraystretch}{1.08}
\caption{Definitions of the eight violation labels used in the human annotations.}
\label{tab:violation-labels}
\begin{tabular}{@{}p{0.15\linewidth}p{\dimexpr0.85\linewidth-2\tabcolsep\relax}@{}}
\toprule
Label       & Definition \\
\midrule
Causality   & An effect occurs without a visible cause, or events occur in an impossible order. \\
Collision   & Objects pass through one another or fail to exhibit an expected response after impact. \\
Contact     & An object floats without support, rests at an incorrect contact point, or fails to maintain visible contact. \\
Deformation & A rigid object warps or melts, or a soft object fails to deform under an interaction that should alter its shape. \\
Fluid       & Pouring, splashing, containment, or other liquid behavior is physically inconsistent. \\
Friction    & Sliding, sticking, or gripping behavior is inconsistent with the visible interaction or stated surface condition. \\
Gravity     & An object floats, falls incorrectly, or otherwise fails to respond plausibly to gravity. \\
Momentum    & An object accelerates or changes motion without a supporting interaction, including apparent creation or loss of energy. \\
\bottomrule
\end{tabular}
\end{table}

These eight human labels are distinct from the seven physical domains used to organize the benchmark and the tool's specialists (Section~\ref{sec:pipeline}). In particular, the human codebook includes both \textbf{Contact} and \textbf{Collision}, whereas the domain taxonomy has a single collision family. We preserve the human labels as recorded and merge Contact into Collision in all analyses. A blank category field indicates only that no label was recorded; it is not independent confirmation that the clip is physically correct.

Some violations may reasonably receive more than one label. For example, an object passing through a supporting surface may involve both Contact and Collision, while unsupported floating may involve both Contact and Gravity. The multi-select interface preserves these overlaps rather than forcing each failure into a single category.

\subsection{Annotator Allocation and Reliability Limits}
\label{app:annotators}

Annotations were produced by members of the research team, mainly graduate students in computer science, rather than by externally recruited participants. Clips were assigned to annotators at random. All observable clips were distributed in batches of 16--17 clips per annotator. Unobservable clips were distributed in batches of 6.

Some observable Cosmos records contain a combined annotator label. Because these records contain only one final set of scores rather than separate ratings attributable to each individual, we treat the combined label as a single annotation source. It does not constitute overlapping independent annotation and cannot be used to estimate agreement between the named annotators. Each clip received one final annotation rather than independent ratings from multiple annotators. Consequently, the dataset does not support an estimate of inter-annotator reliability. 

\section{PhysicsLENS Pipeline Details}
\label{app:pipeline-details}

This appendix describes the pipeline as it was run for Section~\ref{sec:results-tool}. Where the interactive tool of Section~\ref{sec:pipeline} does more, we say so.

\paragraph{Screening.} Stage~1 measures motion over the first 5\,s of each video, sampled at 8 frames per second in grayscale with the longer side resized to 320 pixels. It computes the mean absolute difference between consecutive frames; dense optical flow \citep{farneback2003two}; camera motion, taken as the median flow vector of each frame pair; object motion, the flow left after subtracting camera motion; and the entropy of object-motion directions over the fastest-moving 10\% of pixels. It also embeds the eight VLM frames (below) with DINOv2~\citep{oquab2024dinov2} and measures how the embedding moves: the step size between frames, how that step size changes, whether successive steps keep their direction, and net displacement relative to path length. Each measurement is reduced to one number per video (a mean, 95th percentile, maximum, or ratio), giving 22 screening signals. No signal is thresholded into a flag. Signals become scores only when evidence is combined, by ranking videos against each other (below). Stage~1 also asks the VLM the physics-error question (Table~\ref{tab:vlm-prompts}). It asks the model to count specific physics errors (objects passing through each other, appearing or vanishing, floating, changing shape impossibly, or moving untouched) and to ignore task success and the amount of motion.

\paragraph{Localization.} Stage~2 follows up to 120 corner points from the first frame through the same frames with Lucas--Kanade tracking~\citep{lucas1981iterative}. It removes camera motion by subtracting each frame's median point velocity, then computes 16 signals from the remaining motion:
the number of points, the fraction still tracked at the end, and the fraction lost away from the frame border, which is how an object vanishing mid-video shows up;
the mean point speed and the 95th percentiles of speed, acceleration and jerk;
seven statistics aimed at single families: the vertical acceleration of downward-moving points and its magnitude (gravity), speed after the sharpest slowdown relative to speed before it (momentum), the rate of direction reversals, the growth and fluctuation of the point cloud's spread (deformation), and centroid travel relative to rotation (friction);
and the time and relative strength of the frame with the largest mean point acceleration.
In the interactive tool, Stage~2 goes further. A VLM names the objects, SAM~3~\citep{carion2026sam} segments and tracks them, Stage-1 flags are grouped into time windows, and each constraint family is ranked with a confidence and a time window. That confidence is the larger of a heuristic prior from the Stage-2 evidence and a VLM triage score. Section~\ref{sec:results-tool} uses only the 16 point-track signals.

\paragraph{VLM verification.} Stage~3 asks each of the ten VLMs three more questions (Table~\ref{tab:vlm-prompts}): task completion, violation type and, on unobservable videos only, adherence to the stated property. All four questions, including the Stage-1 physics-error question, show eight frames spaced evenly from the first to the last frame, with the longer side at most 512 pixels (Llama-4-Scout receives four frames). Every question states the task. Only the hidden-property question states the property, and it also gives the scenario's expected outcome. The violation question lists eight failure descriptions and a ``no physics problem'' option. The eight descriptions are one for each family in Table~\ref{tab:vlm-prompts} plus object permanence (an object appearing or vanishing). The options are shuffled per video with a fixed seed, so a preference for the first-listed option does not favor any family. These options are descriptions, not the human labels. There is no Contact option, so Contact is merged into Collision when scoring, and permanence has no human label.

\begin{table}[t]
  \centering
  \footnotesize
  \caption{VLM questions, verbatim (line breaks omitted). Each prompt follows the frames and the header; $\langle\cdot\rangle$ marks fields filled per video. The violation options appear as A)--I) in a per-video shuffled order; family names in parentheses are not shown to the model.}
  \label{tab:vlm-prompts}
  \begin{tabular}{@{}>{\raggedright\arraybackslash}p{0.2\linewidth}>{\raggedright\arraybackslash}p{0.75\linewidth}@{}}
    \toprule
    Question & Prompt \\
    \midrule
    Header & Look at these $\langle n\rangle$ frames, sampled in order from a video of a robot. The task: ``$\langle$task$\rangle$''. \\
    Physics error (plausibility) & Ignore whether the action succeeds and ignore how much motion there is. Count only PHYSICS ERRORS: objects passing through each other, appearing or vanishing, floating without support, changing shape impossibly, or moving with nothing touching them. 1 = many clear physics errors, 2 = some errors, 3 = minor oddities, 4 = no physics errors. Answer with exactly one digit, 1 to 4. \\
    Standard plausibility (baseline) & Rate how PHYSICALLY PLAUSIBLE this video is --- do objects, contact, motion, liquids and deformation behave as they would in the real world? 1 = clearly impossible physics, 2 = noticeable physics errors, 3 = minor oddities, 4 = fully plausible. Answer with exactly one digit, 1 to 4. \\
    Task completion & Did the robot actually COMPLETE the task shown? Answer with exactly one digit: 1 = no, 2 = yes. \\
    Hidden property (unobservable only) & Hidden property of this scene (not visible in the frames): $\langle$property$\rangle$ --- $\langle$value$\rangle$. If the video respects it, this should happen: $\langle$expected outcome$\rangle$ Rate how well the video FOLLOWS that hidden property. 1 = contradicts it, 2 = mostly contradicts, 3 = mostly follows, 4 = clearly follows. Answer with exactly one digit, 1 to 4. \\
    Violation type & (Header: Look at these $\langle n\rangle$ frames, sampled in order from a video of: ``$\langle$task$\rangle$''.) Which ONE of these best describes the main physics problem in this video? $\langle$options$\rangle$ Answer with exactly one letter. \\
    \midrule
    \multicolumn{2}{@{}l}{\emph{Violation options}} \\
    (Collision) & Two things overlap or pass through each other instead of meeting at their surfaces, or the gripper grasps without closing on the object \\
    (Deformation) & A rigid object changes its shape, length or thickness \\
    (Causality) & An object moves or changes on its own, with nothing visibly touching it \\
    (Gravity) & Something hangs in the air, floats, or fails to fall when nothing is holding it \\
    (Momentum) & Something speeds up after being released or struck, or stops dead for no reason \\
    (Permanence) & An object appears from nowhere or vanishes \\
    (Fluid) & Liquid appears, vanishes, or holds an impossible rigid shape \\
    (Friction) & Something slides when it should grip, or keeps sliding with nothing pushing it \\
    (None) & No physics problem --- everything behaves as it should \\
    \bottomrule
  \end{tabular}
\end{table}

Each answer is read from the probabilities of the VLM's first output token. We sum the probability of each allowed answer over its surface forms (e.g.\ ``3'' and ``\,3''), renormalize over the allowed answers, and take the expected rating for the 1--4 questions and $P(\mathrm{yes})$ for completion. Violation detection uses $1-P(\mathrm{none})$, and family attribution uses $P(\text{family})$. A missing answer is replaced by that VLM's mean score. For the spread in Table~\ref{tab:backbones-obs-unobs}, two further runs shift the eight frames by one third and two thirds of the spacing between them. Section~\ref{sec:results-tool} does not use the tool's specialist modules, which fit trajectories and check individual events with a VLM (Section~\ref{sec:pipeline}). There, Stage~3 consists only of these questions.

\paragraph{Combining evidence.} For each question, each VLM's scores are converted to ranks over the 439 videos, scaled to $[0,1]$, and averaged over the ten VLMs, so that no model's calibration dominates. The 38 Stage-1--2 signals are reduced to one signal score per question. On the training folds, we keep the five signals with the largest absolute Spearman correlation with the human label (the 1--4 rating for plausibility and adherence, the binary label otherwise), flip the sign of those that correlate negatively, and average their ranks. The final score is $w\,r_{\mathrm{VLM}} + (1-w)\,r_{\mathrm{sig}}$, where $r$ denotes a rank and $w \in \{0, 0.1, \dots, 1\}$ maximizes AUC on the training folds. The training-fold signal scores used to choose $w$ are themselves computed out of fold, and test videos are ranked against the training videos. Signal selection and $w$ are refit for every question and fold, using five-fold cross-validation that keeps every version of a scenario (both observability conditions and all four generators) in the same fold. Family attribution is fitted on violating videos only, and adherence on unobservable videos only.

\paragraph{Reporting and routing.} Stage~4 writes a JSON report for each video from the shared evidence store. The report holds a chronological timeline of findings, each with its time window, object, violation type, confidence and the specialist's explanation, together with the Stage-2 family ranking and each tool's severity. An optional text LLM summarizes the report in prose. Stage~4 adds no score of its own. In the interactive tool, routing happens between Stages~2 and~3: only the specialists that Stage~2 ranks highest (three by default) run on a video. Section~\ref{sec:results-tool} applies no routing, so every signal and every question is computed for all 439 videos.

\paragraph{VideoPhy-2 check.}
\label{app:videophy}
To test whether the physics-error question helps outside robot videos, we compare it with the standard question on VideoPhy-2~\citep{bansal2026videophy}. We use 1,200 videos from its test split, sampled evenly across the seven generators and the five physical-commonsense ratings. We count a video as plausible when its human physical-commonsense score is at least 4 and as implausible when it is at most 2, and drop the 274 videos rated 3, which leaves 926 videos (541 plausible, 385 implausible). We ask Qwen3-VL-32B, Qwen3-VL-8B and Qwen2.5-VL-7B both questions with the same frame sampling and answer scoring as above, and compare the two AUCs with a paired DeLong test. Qwen3-VL-32B improves from 0.645 to 0.720 ($p<10^{-6}$), Qwen3-VL-8B from 0.666 to 0.708 ($p=0.006$), and Qwen2.5-VL-7B from 0.581 to 0.655 ($p<0.001$).

\section*{Reproducibility Statement}
\label{app:reproducibility}
Dataset construction, prompt schema, generation settings, and the domain taxonomy are described in Appendix~\ref{app:dataset-details}.
The human annotation protocol and rubric are detailed in Appendix~\ref{app:annotation-details}.

The dataset is publicly available on Hugging Face: \url{https://huggingface.co/datasets/swiftrando/PhysicsLENS}.

The code for the PhysicsLENS evaluation and diagnostic tool is available at
\url{https://github.com/Isaiah-Milkey/PhysicsLENS-benchmark}.

\end{document}